\documentclass{bmvc2k}

\usepackage{booktabs}  
\usepackage{amsmath}
\usepackage{amssymb}
\usepackage{multirow}  
\usepackage[most]{tcolorbox}  
\usepackage{tablefootnote}
\usepackage{enumitem}  
\usepackage{subcaption}
\usepackage[font=small]{caption}
\usepackage[multiple]{footmisc}  
\usepackage[nohyperlinks]{acronym}
\acrodef{cnn}[CNN]{convolutional neural network}
\acrodef{wrn}[WRN]{wide residual networks}
\acrodef{at}[AT]{adversarial training}
\acrodef{pp}[pp]{percentage points}
\acrodef{nlp}[NLP]{natural language processing}
\acrodef{rnn}[RNN]{recurrent neural network}
\acrodef{sota}[SOTA]{state-of-the-art}
\acrodef{ra}[Ra]{Robust architecture}
\acrodef{wd}[WD]{width-depth}
\acrodef{nas}[NAS]{neural architecture search}
\acrodef{pgd}[PGD]{projected gradient descent}
\acrodef{sat}[SAT]{standard adversarial training}
\acrodef{swin}[Swin Transformer]{shifted windows Transformer}
\acrodef{xcit}[XCiT]{cross-covariance image Transformers}
\acrodef{deit}[DeiT]{data-efficient image Transformers}
\acrodef{vit}[ViT]{vision Transformer}
\acrodef{edm}[EDM]{elucidating diffusion model}
\acrodef{aa}[AA]{AutoAttack}
\acrodef{se}[SE]{squeeze and excitation}
\acrodef{bn}[BN]{batch normalization}
\acrodef{relu}[ReLU]{rectified linear unit}
\acrodef{prelu}[PReLU]{parametric \acs{relu}}
\acrodef{gelu}[GELU]{gaussian error linear unit}
\acrodef{silu}[SiLU]{sigmoid linear unit}
\acrodef{psilu}[PSiLU]{parametric \acs{silu}}
\acrodef{pssilu}[PSSiLU]{parametric shifted \acs{silu}}
\acrodef{mlp}[MLP]{multilayer perceptron}
\acrodef{edf}[EDF]{error distribution function}
\def\eg{\emph{e.g}\bmvaOneDot}

\def\etal{\emph{et al}\bmvaOneDot}

\DeclareMathOperator*{\argmin}{argmin}

\mathchardef\mh="2D

\definecolor{myblue}{RGB}{54,144,192}
\definecolor{mypurple}{RGB}{222,119,174}
\definecolor{mygreen}{RGB}{77,146,33}
\definecolor{mygray}{RGB}{150,150,150}
\definecolor{mylightgray}{RGB}{217,217,217}
\tcbset{on line, 
        boxsep=2pt, left=0pt,right=0pt,top=0pt,bottom=0pt,
        colframe=white,colback=mylightgray,  
        highlight math style={enhanced}
}

\newcommand*{\hlbox}[1]{\tcbox{#1}}
\newcommand*{\tabindent}{\hspace{3mm}}

\title{Robust Principles: Architectural Design Principles for Adversarially Robust CNNs}

\addauthor{ShengYun Peng}{https://shengyun-peng.github.io/}{1}
\addauthor{Weilin Xu}{https://xuweilin.org/}{2}
\addauthor{Cory Cornelius}{https://dxoig.mn/}{2}
\addauthor{Matthew Hull}{https://matthewdhull.github.io}{1}
\addauthor{Kevin Li}{https://www.kevinyli.com/}{1}
\addauthor{Rahul Duggal}{http://www.rahulduggal.com/}{1}
\addauthor{Mansi Phute}{mphute6@gatech.edu}{1}
\addauthor{Jason Martin}{jason.martin@intel.com}{2}
\addauthor{Duen Horng Chau}{https://faculty.cc.gatech.edu/~dchau}{1}

\addinstitution{
 Georgia Institute of Technology \\
 Atlanta, GA, USA
}
\addinstitution{
 Intel Corporation \\
 Hillsboro, OR, USA
}

\runninghead{S. Peng et al.}{Robust Architectural Design Principles}

\begin{document}

\maketitle

\begin{abstract}
We aim to unify existing works' diverging opinions on how architectural components affect the adversarial robustness of \acsp{cnn}. 
To achieve our goal, we synthesize a suite of generalizable robust architectural design principles: (a) optimal range for \textit{depth} and \textit{width} configurations, (b) preferring \textit{convolutional} over \textit{patchify} stem stage, and (c) robust residual block design by adopting \acl{se} blocks, and non-parametric smooth activation functions.
Through extensive experiments across a wide spectrum of \textit{dataset scales}, \textit{\acl{at} methods}, \textit{model parameters}, and \textit{network design spaces}, our principles consistently and markedly improve \acl{aa} accuracy: 1--3 \ac{pp} on CIFAR-10 and CIFAR-100, and 4--9 \ac{pp} on ImageNet. The code is publicly available at \href{https://github.com/poloclub/robust-principles}{https://github.com/poloclub/robust-principles}.
\end{abstract}
\section{Introduction}
\label{sec:intro}

\Acp{cnn} and Transformers are staples in computer vision research~\cite{he2016deep,dosovitskiy2020image}, but they are vulnerable to adversarial attacks~\cite{szegedy2013intriguing,goodfellow2014explaining,mo2022adversarial}, and \ac{at} is the most effective way to improve their robustness~\cite{madry2018towards}.
While \acp{cnn} and Transformers have comparable clean accuracy~\cite{liu2022convnet,woo2023convnext}, there is a disparity between the two families in robustness research.
\ac{at} has robustified a wide range of Transformers \cite{mo2022adversarial,debenedetti2022light}.
Yet, for \acp{cnn}, \ac{wrn}~\cite{zagoruyko2016wide}, proposed back in 2016, is still the most studied architecture 
in \ac{at} research~\cite{wang2023better,huang2021exploring}.

Although a variety of instructions have been proposed to enhance the robustness of \ac{wrn}, there are still conflicting opinions on how architectural components impact the overall robustness.
For example, recent research~\cite{huang2021exploring,huang2022revisiting}
explored how scaling depth and width might be correlated with improved adversarial robustness;
however, Mok \etal~\cite{mok2021advrush} suggested such a relationship was unclear.
As another example, for micro block design, Xie \etal~\cite{xie2020smooth} and Bai \etal~\cite{bai2021transformers} found smooth activation functions improved robustness, but Huang \etal~\cite{huang2022revisiting} suggested that the performance was dependent on \ac{at} settings.
Furthermore, as most existing robust architecture research experimented only on the small-scale CIFAR-10~\cite{mok2021advrush,mo2022adversarial,huang2022revisiting,huang2021exploring}, it was unclear whether such findings might generalize to large-scale datasets, \eg, ImageNet~\cite{deng2009imagenet}.
For example, while prior work suggested that a straightforward application of \ac{se} hurts robustness on CIFAR~\cite{huang2022revisiting}, we find such a modification, in fact, \textit{improves} robustness on ImageNet (Sec.~\ref{sec:se}).
Similarly, while previous work suggested that parametric activation functions are more robust than non-parametric versions on CIFAR~\cite{dai2022parameterizing}, 
we find the reverse holds true on ImageNet (Sec.~\ref{sec:activation}). 
The divergent observations provide strong motivation for us to seek a suite of unified generalizable design principles that can enhance the adversarial robustness of \acp{cnn}. 

To accomplish our goal, we survey and distill four key architectural components underpinning \ac{sota} \acp{cnn} and Transformers that boost adversarial robustness. 
These components span a network's macro and micro designs: \textit{depth and width}, \textit{stem stage}, \textit{\ac{se} block}, and \textit{activation}.
Different from previous research on robust architectures that solely relies on the small-scale CIFAR-10, we explore the \ac{at} of all four components on the large-scale ImageNet and evaluate on its full validation set. 
Through extensive experiments over the wide spectrum of \textit{dataset scales}, \textit{\ac{at} methods}, \textit{model parameters}, and \textit{network design spaces}, we make the following key contributions:

\begin{figure}[t]
  \centering
   \includegraphics[width=0.68\linewidth]{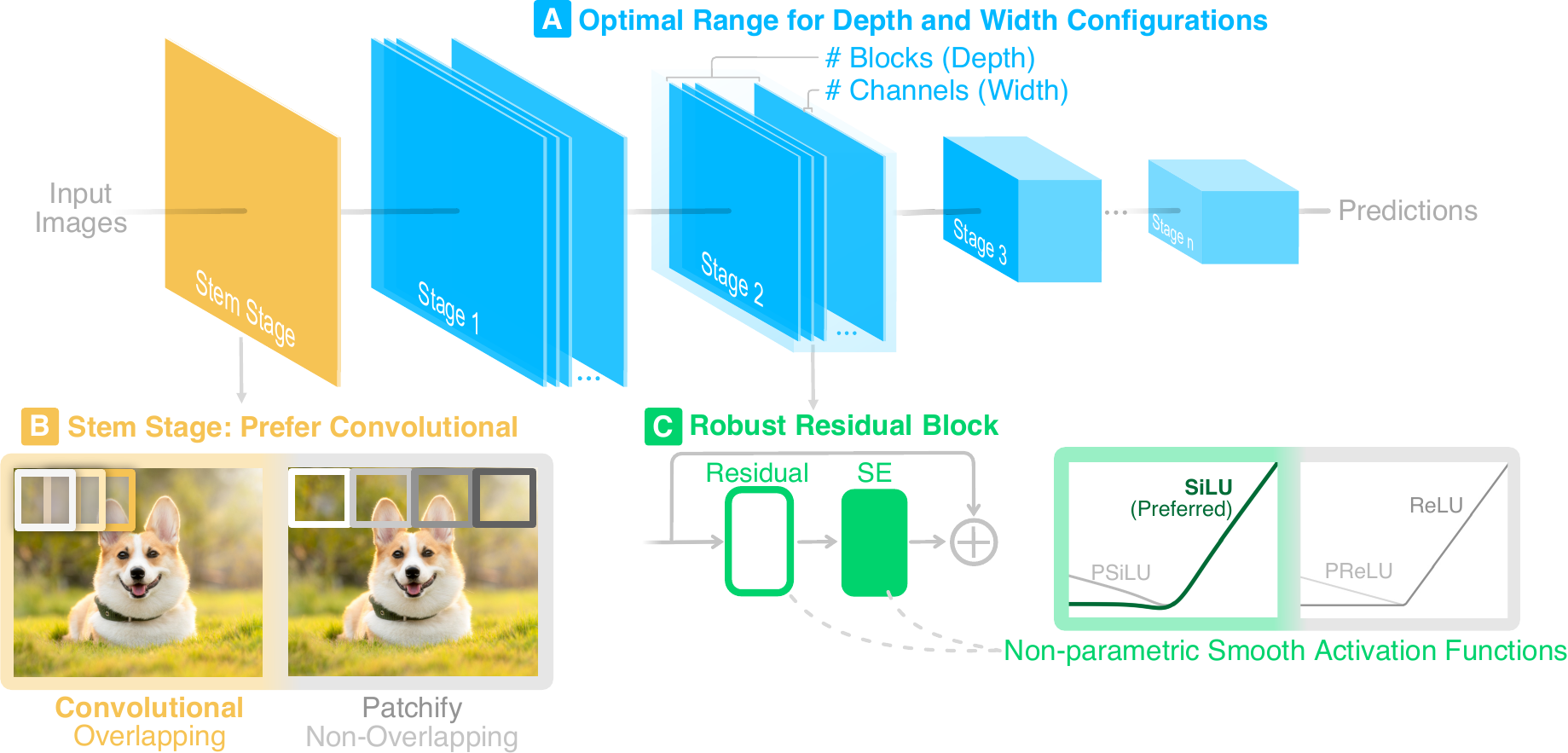}
   \caption{
   We synthesize a suite of generalizable architectural design principles to robustify \acp{cnn}, spanning a network's macro and micro designs: 
   (A) optimal range for depth and width configurations, 
   (B) preferring \textit{convolutional} over \textit{patchify} stem stage, and 
   (C) robust residual block design by adopting \acl{se} blocks, and non-parametric smooth activation functions.
   The principles consistently and markedly improve \acl{aa} accuracy for CIFAR-10, CIFAR-100, and ImageNet over the wide spectrum of \ac{at} methods, model parameters, and network design spaces.
   }
   \label{fig:skeleton}
\end{figure}

\begin{enumerate}[leftmargin=*,topsep=0pt]
\itemsep-0.1em 
\item \textbf{A suite of generalizable robust architectural design principles for \acp{cnn}} (Fig.~\ref{fig:skeleton}): 

\textbf{(a)} Optimal Range for Depth and Width Configurations.  
Despite the popularity of 4-stage residual networks on ImageNet, existing exploration has been constrained to 3-stage designs~\cite{huang2021exploring,huang2022revisiting}.
We discover a flexible depth and width scaling rule that does not place a restriction on the total number of stages, and we verify its generalizability and optimality through extensive experiments. (Sec. \ref{sec:depth_width})

\textbf{(b)} \textit{Convolutional} over \textit{Patchify} Stem Stage. 
\textit{Convolutional} stem and \textit{patchify} stem are commonly used in \acp{cnn} and Transformers to downsample input images due to the redundancy inherent in natural images.
\textit{Convolutional} stem uses overlapped convolution kernel to slide the input image, while \textit{patchify} stem patchifies the input image into $p \times p$ non-overlapping patches. 
We discover that convolutional stem, especially with a postponed downsampling design, outperforms patchify stem due to its less-aggressive stride-two downsampling and overlapped convolution kernels. (Sec. \ref{sec:stem_stage})

\textbf{(c)} Robust Residual Block Design. 
Our investigations on how \ac{se} block and activations affect robustness present new findings on ImageNet that differ from previous research on CIFAR.
Through a hyperparameter sweep, we find the reduction ratio $r$ in \ac{se} block is negatively correlated with robustness, a new discovery not previously reported as prior work was based on a fixed $r$ \cite{huang2022revisiting}.
We also confirm that non-parametric smooth activations consistently improve robustness on CIFAR-10, CIFAR-100, and ImageNet. (Sec. \ref{sec:robust_block})

\item \textbf{Consistent and marked improvements on adversarial robustness.}

We verify the generalization of the three design principles across 
a wide spectrum of \textit{dataset scales} (CIFAR-10, CIFAR-100~\cite{krizhevsky2009learning}, ImageNet~\cite{deng2009imagenet}), \textit{\ac{at} methods} (\ac{sat}~\cite{madry2018towards}, TRADES~\cite{zhang2019theoretically}, Fast-\ac{at}~\cite{wong2020fast}, MART~\cite{wang2020improving}, and diffusion-augmented \ac{at}~\cite{wang2023better}), \textit{model parameters} (from 26M to 267M), and \textit{network design spaces} (variants of \ac{wrn}~\cite{zagoruyko2016wide} and ResNet~\cite{he2016deep}). 
Our experiments demonstrate that all design principles consistently and markedly improve \ac{aa} accuracy by 1--3 \acf{pp} on CIFAR-10 and CIFAR-100 --- boosting even the \ac{sota} diffusion-augmented \ac{at}~\cite{wang2023better} by such amounts --- and 4--9 \ac{pp} on ImageNet. 
In particular, our robustified \ac{wrn}-70-16 boosts the \ac{aa} accuracy by 1.31 \ac{pp} (65.02\% $\rightarrow$ 66.33\%) on CIFAR-10, and 0.96 \ac{pp} (37.77\% $\rightarrow$ 38.73\%) on CIFAR-100. 
On ImageNet, the \ac{aa} accuracy is boosted by 6.48 \ac{pp} (39.78\% $\rightarrow$ 46.26\%) through robustified ResNet-101, and 6.94 \ac{pp} (42.00\% $\rightarrow$ 48.94\%) through robustified \ac{wrn}-101-2.
Our findings unify prior works' diverging opinions on how architectural components affect robustness and highlight the benefits of exploring intrinsically robust architectural components.

\end{enumerate}

\section{Related Work}
\label{sec:related_work}

\noindent \textbf{\Acf{at}.} 
\Ac{at} is an effective approach to defending against adversarial attacks~\cite{goodfellow2014explaining}.
Madry \etal~\cite{madry2018towards} formulated \ac{sat} as a min-max optimization framework.
Given dataset samples $(x_i, y_i)$, network $f_\theta$ and loss function $\mathcal{L}$, the optimization is formulated as:
\begin{equation}
    \argmin_\theta \mathbb{E}_{(x_i, y_i) \sim \mathbb{D}} \left[ \max_{x'} \mathcal{L} \left( f_\theta, x', y \right) \right],  
\end{equation}
The inner adversarial example $x'$ aims to find the perturbation of a given data point $x$ that achieves a high loss and is generated on the fly during the training process.
Since then, multiple variants of \ac{sat} are proposed~\cite{wu2020adversarial,dong2020adversarial, wang2021convergence,bai2021improving,ding2018mma}. 
Our architectural research is complementary to these works on \ac{at} methods. 
We train on diverse \ac{at} methods, \eg, \ac{sat}~\cite{madry2018towards}, Fast-\ac{at}~\cite{wong2020fast}, TRADES~\cite{zhang2019theoretically}, MART~\cite{wang2020improving}, and diffusion-augmented \ac{at}~\cite{wang2023better} to verify that our design principles unanimously improve robustness agnostic to the training recipe. 

\noindent \textbf{Robust Architectures.} 
Here we provide a brief overview of related research on robust architectures, and the extended version is in Sec.~\ref{sec:appx-related-work} in supplementary materials. 
Our key advancement over existing works is a suite of three robust architectural design principles verified on diverse dataset scales, \ac{at} methods, model parameters, and design spaces.  
A few research studied the impact of architectural designs on adversarial robustness~\cite{xie2019intriguing,guo2020meets,singh2023revisiting,peng2023robarch}. 
For macro network design, Huang \etal~\cite{huang2021exploring} and Huang \etal~\cite{huang2022revisiting} led the exploration of the correlations between robustness improvement and scaling depth and width, but Mok \etal~\cite{mok2021advrush} suggested such a relationship was unclear. 
For micro block design, smooth~\cite{xie2020smooth,bai2021transformers} and parameterized~\cite{dai2022parameterizing} activation functions can largely improve robustness on CIFAR, but Huang \etal~\cite{huang2022revisiting} found the performance was dependent on \ac{at} settings. 
There was no clear consensus on how architectural components affect adversarial robustness. 
More importantly, most conclusions are drawn on CIFAR with \ac{wrn}'s basic block design. 
It was unclear whether such existing robust architectures generalize to large-scale datasets or other network design spaces.
Our work provides conclusive evidence that unifies prior works' diverging opinions on how architectural components affect robustness. 

\section{Preliminaries}
\label{sec:preliminaries}

This section describes the setups for comparing \acp{cnn} and Transformers in terms of architectural design space, training techniques, and adversarial attacks. 

\noindent \textbf{Architectural Design Skeleton.} 
Fig.~\ref{fig:skeleton} provides the \ac{cnn} skeleton that supports modifications of different architectural components in our study. 
Specifically, the skeleton consists of a stem stage, $n$ body stages, and a classification head.
A typical body stage has multiple residual blocks, and the block type is either basic (two $3 \times 3$ convolutions) or bottleneck ($1 \times 1$, $3 \times 3$, and $1 \times 1$ convolutions). 
For stage $i$, denote \textit{depth} $D_i$ as number of blocks, and \textit{width} $W_i$ as channels in $3 \times 3$ convolution. 
The downsampling factor is 1 in the first block of stage 1, and 2 in the first block of stage 2 to $n$. 
Unless otherwise specified, the default operations in each block are \ac{relu} and \ac{bn}. 

\noindent \textbf{Training Techniques.} 
We use five \ac{at} recipes: Fast-\ac{at}~\cite{wong2020fast}, \ac{sat}~\cite{madry2018towards}, TRADES~\cite{zhang2019theoretically}, MART~\cite{wang2020improving}, and diffusion-augmented \ac{at}~\cite{wang2023better}. 
The architectural exploration is conducted on ImageNet~\cite{deng2009imagenet}, and we apply the same training method so that the performance difference between models can only be attributed to the difference in architectures.
Due to the size of ImageNet and the slow speed of \ac{at}, we use Fast-\ac{at}~\cite{wong2020fast} with a cyclic learning rate~\cite{smith2017cyclical} and mixed-precision arithmetic~\cite{micikevicius2017mixed}. 
After finalizing the architectural design principles, we train all models with diverse \ac{at} methods on CIFAR-10, CIFAR-100~\cite{krizhevsky2009learning} and ImageNet~\cite{deng2009imagenet} to verify that our designs can consistently improve robustness regardless of the training recipe. 

\noindent \textbf{Adversarial Attacks.}
\Ac{pgd}~\cite{madry2018towards} and \ac{aa}~\cite{croce2020reliable} are used to evaluate adversarial robustness.
\ac{pgd} is a white-box attack with complete access to network architectures and parameters. 
For \ac{pgd}, we provide a comprehensive investigation on the full ImageNet validation set with attack budgets $\epsilon \in \{2, 4, 8 \} / 255$ and max steps $i = \{10, 50, 100\}$, denoted as $\ac{pgd}^i \mh \epsilon$. 
\ac{aa} is an ensemble of one black-box and three white-box attacks. 
The attack budget is $\epsilon = 4/255$ on ImageNet~\cite{croce2020robustbench}, and $\epsilon = 8/255$ on CIFAR-10 and CIFAR-100~\cite{huang2021exploring,huang2022revisiting} for \ac{aa}. 
All attacks are $\ell_\infty$ bounded. 
When exploring individual architectural components, we use 10-step \ac{pgd} ( $\ac{pgd}^{10} \mh \epsilon$) for fast evaluations.

\section{Robust Architectural Design Principles}
\label{sec:robust_arch_design}

Our strategy to explore the four robust architectural components is through comparing \ac{sota} \acp{cnn} and Transformers.
These components span a network's macro and micro designs: depth and width (Sec. \ref{sec:depth_width}), stem stage (Sec. \ref{sec:stem_stage}), \acl{se} block (Sec. \ref{sec:se}), and activation (Sec. \ref{sec:activation}), as shown in Fig.~\ref{fig:skeleton}.
To exclude the effect of model complexity~\cite{madry2018towards} and provide a fair comparison~\cite{bai2021transformers}, we focus our investigation on the regime of ResNet-50 with ${\sim} 26$M parameters in this section. 
In the next section (Sec. \ref{sec:experiments}), we verify the generalization of these principles through extensive experiments over the wide spectrum of dataset scales, \ac{at} methods, parameter budgets, and network design spaces. 

\subsection{Optimal Range for Depth and Width Configurations}
\label{sec:depth_width}

Macro network design involves the distribution of depth and width in each stage. 
\Ac{swin}~\cite{liu2021swin} regulates the stage compute ratio as $1:1:3:1$ or $1:1:9:1$. 
On the \ac{cnn} side, ConvNeXt~\cite{liu2022convnet} reshuffles depths in all stages according to \ac{swin} and finds this design improves clean accuracy. 
RegNet~\cite{radosavovic2020designing} introduces a linear parameterization to assign network depth and width. 
As network depth and width are competing for resources when the parameter budget is fixed, 
it is important to study the impact of depth and width on adversarial robustness.
We draw inspiration from prior robustness research to develop a more generalized scaling rule. 
Huang \etal~\cite{huang2021exploring} found reducing depth or width at the last stage of \ac{wrn} reduces the Lipschitz constant, thus improving robustness. 
Huang \etal~\cite{huang2022revisiting} proposed a fixed scaling ratio for \ac{wrn}~\cite{zagoruyko2016wide} on CIFAR-10. 

\begin{figure}[!htbp]
\centering
    \begin{subfigure}{0.52\linewidth}
    \centering
        \includegraphics[height=1.6in]{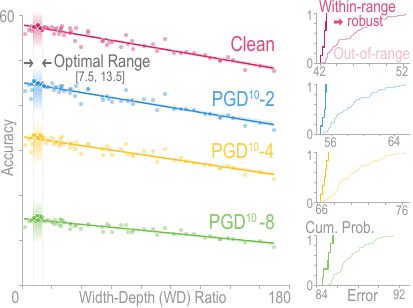}
        \caption{Depth and Width Configurations}
        \label{fig:dw-ratio}
    \end{subfigure}
    \begin{subfigure}{0.45\linewidth}
        \includegraphics[height=1.6in]{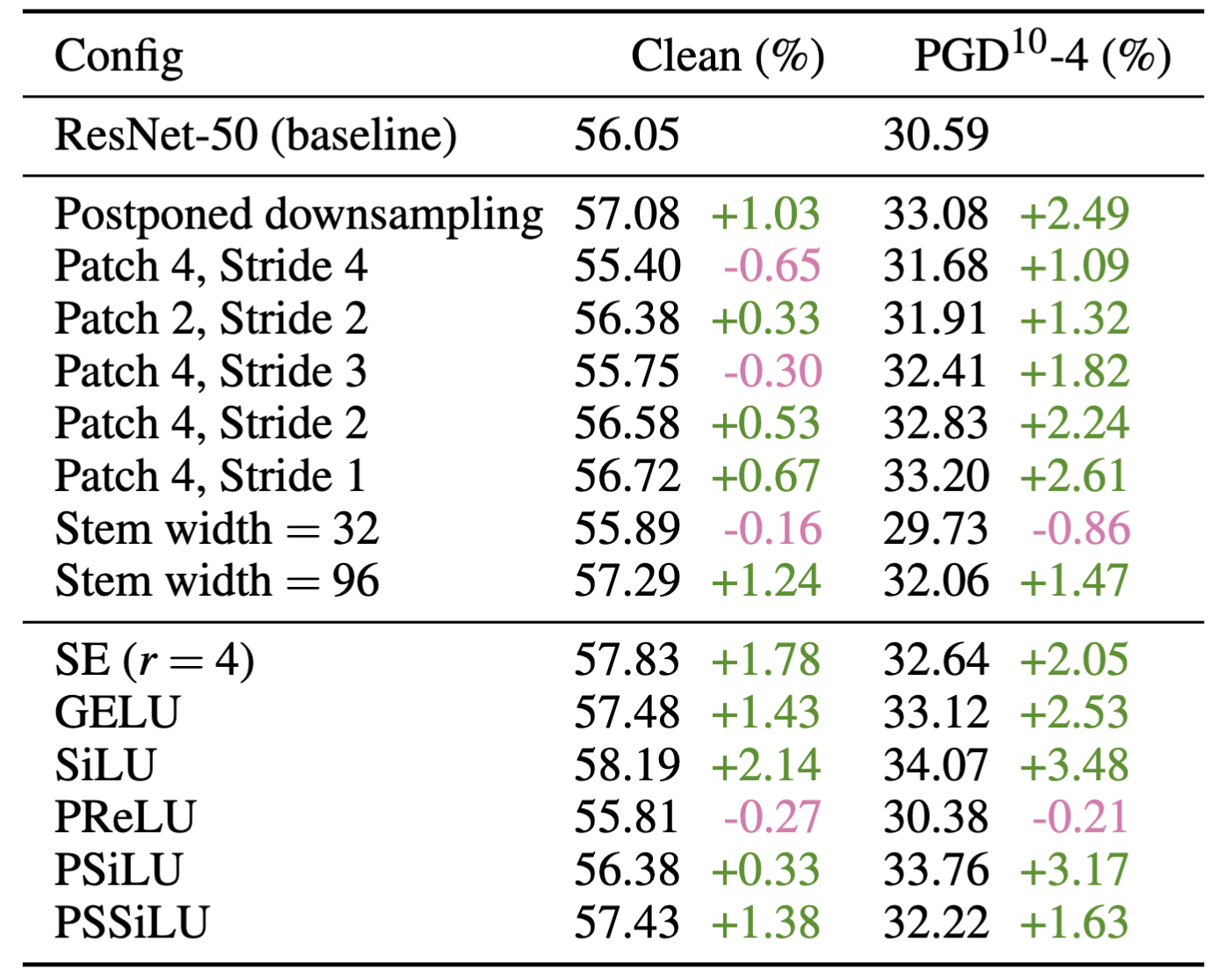}
        \caption{
        Stem Stage and Residual Block Designs
        }
        \label{fig:other-components}
    \end{subfigure}
\caption{
\textbf{(a)} Clean and \ac{pgd} accuracies are negatively correlated with the \ac{wd} ratio. 
Each dot is a configuration of \#stage, depth, and width. 
Intersecting each attack budget's top 10\% most accurate configurations, we find the optimal range of the \ac{wd} ratio is $[7.5, 13.5]$ (with color backgrounds) and verify the significance by the \acp{edf} of models within range (steeper dark lines) and outside of range (gentle light lines).  
\textbf{(b)} Performance of different configurations in stem stage and residual blocks. 
Differences are compared with baseline ResNet-50. 
All models trained with Fast-\ac{at}~\cite{wong2020fast} and evaluated on full ImageNet validation set.
Different \ac{pgd} attack budgets show a similar accuracy trend, and the full results are shown in supplementary material Sec.~\ref{sec:appx-other-components}.
}
\label{fig:components}
\end{figure}

Existing explorations are constrained to 3-stage networks, whereas 4-stage residual networks are more commonly used on ImageNet. 
Thus, instead of studying a fixed depth and width configuration in each stage, we aim to provide a flexible compound scaling rule of the relationship between robustness and total depths and widths. 
Define the \ac{wd} ratio of a $n$-stage network as the average of comparing $W_i$ to $D_i$ in stage $i$: 
\begin{equation}
    \text{WD ratio} = \frac{1}{n - 1}\sum_{i = 1}^{n - 1}\frac{W_i}{D_i}
\end{equation}
The last stage is excluded since reducing its capacity improves robustness. 
\ac{wd} ratio has three parameters: total stages $n$, depth $D_i$, and width $W_i$. 
To obtain valid models, we randomly sample from $n \in \{3, 4, 5, 6 \}$, $D_i \leq 60$, and $W_i \leq 1000$. 
Fig.~\ref{fig:dw-ratio}-left shows the \ac{at} results of all samples. 
All clean and \ac{pgd} accuracies show negative correlations with the \ac{wd} ratio. 
Note when the \ac{wd} ratio is close to 0, the \ac{at} is unstable and also leads to inferior robustness. 
We intersect the \ac{wd} ratio of top 10\% networks from each attack budget and find the optimal range of \ac{wd} ratio is $[7.5, 13.5]$. 
We use \ac{edf} (Fig.~\ref{fig:dw-ratio}-right) to provide the characteristics of models from within and outside of the optimal range.
A marked robustness gain is achieved by simply re-distributing depths and widths to satisfy the optimal range. 
Compare to ResNet-50's WD ratio of 32, a robust network has lower \ac{wd} ratio, which echoes previous research that deep and narrow networks are better than shallow and wide networks~\cite{huang2022revisiting}. 
Furthermore, our \ac{wd} ratio is not limited to 3-stage networks as ResNet-50 shown here already has 4 stages. 
Sec. \ref{sec:experiments} shows generalization to other networks.

\subsection{\textit{Convolutional} over \textit{Patchify} stem stage}
\label{sec:stem_stage}

As a preprocessor, a common stem stage ``aggressively'' downsamples input images due to the redundancy inherent in natural images. 
The stem stage in ResNet-50~\cite{he2016deep} consists of a stride-two $7 \times 7$ convolution and a stride-two max-pooling. 
Built on ResNet-50, RegNet~\cite{radosavovic2020designing} replaces the max-pooling with a stride-two residual shortcut in the first block of the first body stage, dubbed \textit{postponed downsampling}.
On the Transformer side, the stem stage patchifies the image into $p \times p$ non-overlapping patches.
It is implemented by a stride-$p$ $p \times p$ convolution, with $p = 14/16$ in \ac{vit}~\cite{dosovitskiy2020image}, and $p = 4$ in \ac{swin}~\cite{liu2021swin}. 
Both \textit{patchify stem} and ResNet-style \textit{convolutional stem} are applicable to \acp{cnn} and Transformers. 
As a pure \ac{cnn}, ConvNeXt~\cite{liu2022convnet} borrowed the patchify stem from Transformer, and Xiao \etal~\cite{xiao2021early} successfully employed the convolutional stem in \ac{vit}. 

Inspired by these findings, we compare how patchify and convolutional stem affect adversarial robustness. 
We set $p = 4$ following \ac{swin} and ConvNeXt for patchify stem (Patch 4, Stride 4) and use postponed downsampling for convolutional stem. 
From the results in Fig.~\ref{fig:other-components}, we observe both designs show better robustness than the baseline ResNet-50, but postponed downsampling is significantly higher than patchify stem. 
There are two differences between these two designs: a smaller stride and the overlapping region between two convolution kernels in the convolutional stem. 
We verify whether these two distinctions are beneficial to robustness. 
First, we reduce the patch size from $4 \times 4$ to $2 \times 2$, and add the stride-two residual shortcut as in postponed downsampling to maintain the total downsampling ratio. 
This modification (Patch 2, Stride 2) improves $\ac{pgd}^{10} \mh 4$ and clean accuracies of $4 \times 4$ patch by 0.23 and 0.98 \ac{pp}. 
Next, we gradually increase the overlapping area between the $4 \times 4$ patch by decreasing the stride from 3 to 1. 
We observe a consistent increment while decreasing the stride, and the $4 \times 4$ patch with the largest overlapping between neighboring patches (Patch 4, Stride 1) performs almost on par with the postponed downsampling. 
Finally, we additionally experiment on different output channel widths since it barely changes total parameters. 
Decreasing the width from 64 (ResNet-50) to 32 lowers the accuracy due to fewer model parameters. 
However, increasing the width from 64 to 96 boosts $\ac{pgd}^{10} \mh 4$ and clean accuracies by 1.47 and 1.24 \ac{pp} with a negligible 0.01M increase in total parameters.
In summary, the convolution stem with postponed downsampling design outperforms patchify stem due to its less aggressive stride-two downsampling and overlapped convolution kernels. 
Besides, widening the output channels significantly improves robustness almost at no cost.

\subsection{Robust Residual Block Design}
\label{sec:robust_block}
\subsubsection{Squeeze \& Excitation}
\label{sec:se}
\Acf{se} block is a simple but effective add-on for the original residual block. 
It was proposed to adaptively recalibrate channel-wise feature responses by modeling interdependencies between channels~\cite{hu2018squeeze}. 
RegNet~\cite{radosavovic2020designing} also verifies the effectiveness of \ac{se} on improving clean accuracy. 
However, Huang \etal~\cite{huang2022revisiting} found that a straightforward application of \ac{se} hurts robustness on CIFAR-10, and added an extra skip connection around the \ac{se}. 
Driven by the discrepancy in clean and adversarial accuracies, we examine the \ac{se} block with ResNet-50. 
Note that the \ac{se} block will introduce additional parameters. 
Thus, we follow RegNet that sets $r = 4$ and appends the \ac{se} block directly after the $3 \times 3$ convolutional layer  since it possesses fewer channels. 
Our experiments on ImageNet show a different picture, where the original \ac{se} block markedly increases all \ac{pgd} and clean accuracies compared to ResNet-50 (Fig.~\ref{fig:other-components}).
We hypothesize that the different effects on CIFAR and ImageNet are caused by the reduction ratio $r$ since Huang \etal only tested on $r = 16$.
To provide a fair comparison, we train \ac{wrn} with a sweep of hyperparameter $r = \{2, 4, 8, 16, 32, 64 \}$. 
Our experiments show that adversarial robustness is negatively correlated with $r$, and when $r \geq 32$, the accuracy is inferior to the baseline \ac{wrn}. 
More details are in supplementary material Sec.~\ref{sec:appx-se}. 
Therefore, \ac{se} block is a unified architecture component that improves both clean and adversarial accuracies, and we set $r = 4$ in all designs that include the \ac{se} block.

\subsubsection{Non-parametric Smooth Activation Functions} 
\label{sec:activation}

Both the \textit{number} and the \textit{function} of activation layers are different between \ac{nlp} and vision architectures. 
In terms of the number, Transformer \cite{dosovitskiy2020image} only has one activation in the \ac{mlp} block. 
In comparison, the activation is normally appended to all convolutions in a \ac{cnn} block~\cite{he2016deep,he2016identity}. 
ConvNeXt~\cite{liu2022convnet} first observed this phenomenon and reduced the total activations from three to one in all blocks, which leads to higher clean accuracy.  
Inspired by ConvNeXt, We combinatorially analyzed all possible locations of the activation in a single block. 
The average $\ac{pgd}^{10} \mh 4$ accuracies are 30.43\%, 29.00\%, and 24.84\% when total activations are set to 3, 2, and 1, which are all inferior to ResNet-50. 
Thus, reducing activation layers is not beneficial to adversarial robustness, and we preserve the activation along with all convolutions. 

In terms of the activation function, it is common practice to use \ac{relu} in \acp{cnn}~\cite{he2016deep} due to its efficiency and simplicity. 
However, advanced Transformers, \eg, BERT~\cite{devlin2018bert}, GPT-2~\cite{radford2019language}, and hierarchical Transformers~\cite{liu2021vision} adopted smoother variants of \ac{relu}, \eg, \ac{gelu}~\cite{hendrycks2016gaussian} and \ac{silu}~\cite{elfwing2018sigmoid}. 
Recent research shows replacing \ac{relu} with its smooth approximations facilitates better gradient updates and leads to higher robustness~\cite{xie2020smooth}.
For example, solely replacing all \acp{relu} with \acp{gelu} significantly improves the \ac{pgd} accuracy of a standard ResNet-50~\cite{bai2021transformers}. 
Dai \etal~\cite{dai2022parameterizing} proposed parametric activation functions by adding learnable parameters to non-parametric versions, which showed mixed performance on CIFAR-10 for different design spaces. 
Compared to \ac{silu}, \ac{psilu} and \ac{pssilu} show higher robustness when applied to \ac{wrn}-28-10, but show degraded performance when applied to ResNet-18.
Therefore, we examine and compare the robustness gain brought by \ac{gelu}, \ac{silu}, \ac{prelu}, \ac{psilu}, and \ac{pssilu}.
We find that: 
(1) non-parametric smooth activations, \eg, \ac{silu} and \ac{gelu} are significantly higher than \ac{relu} (ResNet-50); and 
(2) for parametric activations, \ac{prelu} performs on par with \ac{relu}, and both \ac{psilu} and \ac{pssilu} are inferior to \ac{silu}.
Synthesizing all the above findings, we recommend non-parametric smooth activations due to consistency and simplicity, thus replacing all \acp{relu} with \acp{silu}. 

\begin{table}[!htbp]
\renewcommand{\arraystretch}{0.92}
\small
  \centering
  \caption{A roadmap to robustify ResNet-50 by cumulatively applying our three design principles. We label the robustified version as \hlbox{\acs{ra}}.
  \hlbox{\acs{ra}}ResNet-50 significantly boosts the clean and adversarial accuracies of ResNet-50.
  Cumulative gains and robustified accuracies highlighted in bold.
  All models are trained with Fast-\ac{at}~\cite{wong2020fast} and evaluated on full ImageNet validation set.
  Each configuration is trained thrice and evaluated thrice with different seeds (9 values in total). The standard deviations are shown in parentheses. 
  }
\scalebox{0.8}{
  \begin{tabular}{@{}l@{\hspace{0mm}}r@{\hspace{2mm}}r@{\hspace{1mm}}l@{\hspace{2mm}}r@{\hspace{1mm}}l@{\hspace{2mm}}r@{\hspace{1mm}}l@{\hspace{2mm}}r@{\hspace{1mm}}l@{}}
  \toprule
  ResNet-50 $\rightarrow$ \textbf{\hlbox{\acs{ra}}ResNet-50} & \#Param. & \multicolumn{2}{c}{Clean (\%)} & \multicolumn{2}{c}{$\ac{pgd}^{10} \mh 2$ (\%)} & \multicolumn{2}{c}{$\ac{pgd}^{10} \mh 4$ (\%)} & \multicolumn{2}{c}{$\ac{pgd}^{10} \mh 8$ (\%)} \\
  \midrule
  ResNet-50 (starting point) & 25.7M & 56.05 \scriptsize{(0.10)} & & 42.81 \scriptsize{(0.22)} & & 30.59 \scriptsize{(0.15)} & & 12.62 \scriptsize{(0.12)} \\ 

  \tabindent + Depth \& Width Config (Sec. \ref{sec:depth_width}) & 25.8M & 57.85 \scriptsize{(0.16)} & \textcolor{mygreen}{+1.80} & 45.90 \scriptsize{(0.26)} & \textcolor{mygreen}{+3.09} & 33.87 \scriptsize{(0.10)} & \textcolor{mygreen}{+3.28} & 15.27 \scriptsize{(0.05)} & \textcolor{mygreen}{+2.65}  \\ 
  
  \tabindent + Convolutional Stem Stage (Sec. \ref{sec:stem_stage}) & 25.9M & 58.00 \scriptsize{(0.45)} & \textcolor{mygreen}{+0.15} & 46.59 \scriptsize{(0.52)} & \textcolor{mygreen}{+0.69} & 34.90 \scriptsize{(0.31)} & \textcolor{mygreen}{+1.03} & 15.85 \scriptsize{(0.28)} & \textcolor{mygreen}{+0.58}  \\
  
  \tabindent + Squeeze and Excitation (Sec. \ref{sec:se}) & 26.2M & 60.22 \scriptsize{(0.28)} & \textcolor{mygreen}{+2.22} & 48.95 \scriptsize{(0.08)} & \textcolor{mygreen}{+2.36} & 36.43 \scriptsize{(0.17)} & \textcolor{mygreen}{+1.53} & 16.43 \scriptsize{(0.05)} & \textcolor{mygreen}{+0.58} \\
  
  \tabindent + Smoother Activation (Sec. \ref{sec:activation}) & 26.2M & \textbf{62.02} \scriptsize{(0.03)} & \textcolor{mygreen}{+1.80} & \textbf{51.47} \scriptsize{(0.22)} & \textcolor{mygreen}{+2.52} & \textbf{39.65} \scriptsize{(0.27)} & \textcolor{mygreen}{+3.22} & \textbf{18.97} \scriptsize{(0.14)} & \textcolor{mygreen}{+2.54}\\
  \midrule
  &  & Total: & \textcolor{mygreen}{\textbf{+5.97}}  &  & \textcolor{mygreen}{\textbf{+8.66}} &  & \textcolor{mygreen}{\textbf{+9.06}} &  & \textcolor{mygreen}{\textbf{+6.35}} \\
  \bottomrule
  \end{tabular}
}
\label{tab:roadmap}
\end{table}
\begin{table}[!htbp]
\renewcommand{\arraystretch}{0.19}
\small
\centering
\caption{Adversarial robustness on CIFAR-10 and CIFAR-100 against \ac{aa} and 20-step \ac{pgd} ($\ac{pgd}^{20}$) with the same maximum perturbation $\ell_\infty, \epsilon = 8/255$. 
Applying our principles leads to a consistent 1--3 \ac{pp} robustness gain across \ac{at} methods, parameter budgets, and design spaces,
boosting even the \ac{sota} ``Diff. 1M'' and ``Diff. 50M'' \ac{at} methods proposed by Wang \etal~\cite{wang2023better}. 
Sec.~\ref{sec:appx-cifar} in supplementary materials provides a systematic comparison with Transformers and \ac{nas}-based architectures.
}
\scalebox{0.69}{
\begin{tabular}{@{}l@{\hspace{2mm}}l@{\hspace{2mm}}l|r@{\hspace{1mm}}l@{\hspace{2mm}}r@{\hspace{1mm}}l@{\hspace{2mm}}r@{\hspace{1mm}}l|r@{\hspace{1mm}}l@{\hspace{2mm}}r@{\hspace{1mm}}l@{\hspace{2mm}}r@{\hspace{1mm}}l@{}}
\toprule
    \multirow{2}{*}{\#Param.} & \multirow{2}{*}{Method} & \multirow{2}{*}{Model} & \multicolumn{6}{c}{CIFAR-10} & \multicolumn{6}{c}{CIFAR-100}\\
   & & & \multicolumn{2}{c}{Clean (\%)} & \multicolumn{2}{c}{\acs{aa} (\%)} & \multicolumn{2}{c}{$\ac{pgd}^{20}$ (\%)} & \multicolumn{2}{c}{Clean (\%)} & \multicolumn{2}{c}{\acs{aa} (\%)} & \multicolumn{2}{c}{$\ac{pgd}^{20}$ (\%)} \\
\midrule
    \multirow{28}{*}{26M} & \multirow{3}{*}{SAT} & ResNet-50 & 84.05 & & 49.97 & & 54.37 & & 55.86 & & 23.78 & & 27.48 \\
    & & \hlbox{\acs{ra}}ResNet-50 & \textbf{84.91} & \textcolor{mygreen}{+0.86} & \textbf{50.94} & \textcolor{mygreen}{+0.97} & \textbf{55.19} & \textcolor{mygreen}{+0.82} & \textbf{56.38} & \textcolor{mygreen}{+0.52} & \textbf{24.99} & \textcolor{mygreen}{+1.21} & \textbf{28.84} & \textcolor{mygreen}{+1.36} \\
\cmidrule{2-15}
    & \multirow{3}{*}{TRADES} & ResNet-50 & 82.26 & & 49.91 & & 54.50 & & 56.00 & & 25.05 & & 29.91 \\
    & & \hlbox{\acs{ra}}ResNet-50 & \textbf{82.80} & \textcolor{mygreen}{+0.54} & \textbf{51.23} & \textcolor{mygreen}{+1.32} & \textbf{55.44} & \textcolor{mygreen}{+0.94} & \textbf{56.29} & \textcolor{mygreen}{+0.29} & \textbf{25.83} & \textcolor{mygreen}{+0.78} & \textbf{31.87} & \textcolor{mygreen}{+1.96} \\
\cmidrule{2-15}
    & \multirow{3}{*}{MART} & ResNet-50 & 77.98 & & 47.17 & & 52.70 & & 53.18 & & 25.35 & & 30.79 \\
    & & \hlbox{\acs{ra}}ResNet-50 & \textbf{79.60} & \textcolor{mygreen}{+1.62} & \textbf{49.19} & \textcolor{mygreen}{+2.02} & \textbf{56.47} & \textcolor{mygreen}{+3.77} & \textbf{53.68} & \textcolor{mygreen}{+0.50} & \textbf{26.97} & \textcolor{mygreen}{+1.62} & \textbf{32.81} & \textcolor{mygreen}{+2.02} \\
\midrule
    \multirow{38}{*}{37M} & \multirow{3}{*}{SAT} & \ac{wrn}-28-10 & 85.44 & & 48.45 & &  53.13 & & \textbf{60.49} & & 23.64 & & 27.47 \\
    & & \hlbox{\acs{ra}}\ac{wrn}-28-10 & \textbf{85.52} & \textcolor{mygreen}{+0.08} & \textbf{51.96} & \textcolor{mygreen}{+3.51} & \textbf{56.22} & \textcolor{mygreen}{+3.09} & 59.09 & \textcolor{mypurple}{-1.40} & \textbf{25.14} & \textcolor{mygreen}{+1.50} & \textbf{29.27} & \textcolor{mygreen}{+1.80}  \\
\cmidrule{2-15}
    & \multirow{3}{*}{TRADES} & \ac{wrn}-28-10 & \textbf{83.86} & & 51.79 & & 55.69 & & 55.21 & & 25.47 & & 29.34 \\
    & & \hlbox{\acs{ra}}\ac{wrn}-28-10 & 83.29 & \textcolor{mypurple}{-0.57} & \textbf{52.10} & \textcolor{mygreen}{+0.31} & \textbf{56.31} & \textcolor{mygreen}{+0.62} & \textbf{55.38} & \textcolor{mygreen}{+0.71} & \textbf{25.68} & \textcolor{mygreen}{+0.21} & \textbf{29.41} & \textcolor{mygreen}{+0.07} \\
\cmidrule{2-15}
    & \multirow{3}{*}{MART} & \ac{wrn}-28-10 & 82.83 & & 50.30 & & 57.00 & & 51.31 & & 25.78 & &  30.06\\
    & & \hlbox{\acs{ra}}\ac{wrn}-28-10 & \textbf{82.85} & \textcolor{mygreen}{+0.02} & \textbf{50.81} & \textcolor{mygreen}{+0.51} & \textbf{57.35} & \textcolor{mygreen}{+0.35} & \textbf{51.61} & \textcolor{mygreen}{+0.30} & \textbf{26.11} & \textcolor{mygreen}{+0.33} & \textbf{30.82} & \textcolor{mygreen}{+0.76} \\
\cmidrule{2-15}
    & \multirow{3}{*}{Diff. 1M} & \ac{wrn}-28-10 & 90.61 & & 61.66 & & 66.43 & & 67.26 & & 34.26 & & 39.29 \\
    & & \hlbox{\acs{ra}}\ac{wrn}-28-10 & \textbf{91.32} & \textcolor{mygreen}{+0.71} & \textbf{65.11} & \textcolor{mygreen}{+3.45} & \textbf{68.93} & \textcolor{mygreen}{+2.50} & \textbf{69.03} & \textcolor{mygreen}{+1.77} & \textbf{37.24} & \textcolor{mygreen}{+2.98} & \textbf{41.59} & \textcolor{mygreen}{+2.30} \\
\midrule
    \multirow{15}{*}{67M} & \multirow{3}{*}{SAT} & \ac{wrn}-34-12 & 85.92 & & 49.35 & & 53.05 & & 59.08 & & 23.69 & & 27.05   \\
    & & \hlbox{\acs{ra}}\ac{wrn}-34-12 & \textbf{86.50} & \textcolor{mygreen}{+0.58} & \textbf{51.78} & \textcolor{mygreen}{+2.43} & \textbf{56.04} & \textcolor{mygreen}{+2.99} & \textbf{59.46} & \textcolor{mygreen}{+0.38} & \textbf{25.18} & \textcolor{mygreen}{+1.49} & \textbf{29.49} & \textcolor{mygreen}{+2.44} \\
\cmidrule{2-15}
    & \multirow{3}{*}{Diff. 1M} & \ac{wrn}-34-12 & 91.11 & & 62.83 & & 67.53 & & 68.40 & & 35.67 & & 40.33 \\
    & & \hlbox{\acs{ra}}\ac{wrn}-34-12 & \textbf{91.75} & \textcolor{mygreen}{+0.64} & \textbf{65.71} & \textcolor{mygreen}{+2.88} & \textbf{69.67} & \textcolor{mygreen}{+2.14} & \textbf{69.75} & \textcolor{mygreen}{+1.35} & \textbf{37.73} & \textcolor{mygreen}{+2.06} & \textbf{42.16} & \textcolor{mygreen}{+1.83} \\
\midrule
    \multirow{29}{*}{267M} & \multirow{3}{*}{SAT} & \ac{wrn}-70-16 & 86.26 & & 50.19 & & 53.74 & & 60.26 & & 23.99 & & 27.05\\
    & & \hlbox{\acs{ra}}\ac{wrn}-70-16  & \textbf{86.72} & \textcolor{mygreen}{+0.46} & \textbf{52.13} & \textcolor{mygreen}{+1.94} & \textbf{56.49} & \textcolor{mygreen}{+2.75} & \textbf{60.42} & \textcolor{mygreen}{+0.16} & \textbf{25.17} & \textcolor{mygreen}{+1.18} & \textbf{29.46} & \textcolor{mygreen}{+2.41}\\
\cmidrule{2-15}
    & \multirow{3}{*}{Diff. 1M} & \ac{wrn}-70-16 & 91.82 & & 65.02 & & 69.10 & & 70.10 & & 37.77  & & 41.95 \\
    & & \hlbox{\acs{ra}}\ac{wrn}-70-16 & \textbf{92.16} & \textcolor{mygreen}{+0.34} & \textbf{66.33} & \textcolor{mygreen}{+1.31} & \textbf{70.37} & \textcolor{mygreen}{+1.27} & \textbf{70.25} & \textcolor{mygreen}{+0.15} & \textbf{38.73} & \textcolor{mygreen}{+0.96} & \textbf{42.61} & \textcolor{mygreen}{+0.66} \\
\cmidrule{2-15}
    & \multirow{3}{*}{Diff. 50M} & \ac{wrn}-70-16 & 93.25 & & 70.69 & & 73.89 & & - & & -  & & - \\
    & & \hlbox{\acs{ra}}\ac{wrn}-70-16 & \textbf{93.27} & \textcolor{mygreen}{+0.02} & \textbf{71.07} & \textcolor{mygreen}{+0.38} & \textbf{75.28} & \textcolor{mygreen}{+1.39} & - &  & - & & - & \\
\bottomrule
\end{tabular}
}
\label{tab:results-cifar}
\end{table}
\section{Adversarial Robustness Evaluation}
\label{sec:experiments}

We have completed the exploration of how individual architectural components affect adversarial robustness. 
It is encouraging to uncover how these components affect robustness, but it is not completely convincing unless the following two questions are addressed: 
1) \textit{Can these components consistently improve robustness when grouped together?} 
2) \textit{Are these design principles generalizable?}
Sec.~\ref{sec:roadmap} provides a roadmap that robustifies a \ac{cnn} with all three design principles, dubbed \ac{ra}.
For a model robustified with our design principles, we tag it with \hlbox{\ac{ra}}. 
Then, we scale model architectures under diverse parameter budgets, extrapolate to the design spaces of ResNet~\cite{he2016deep} and \ac{wrn}~\cite{zagoruyko2016wide}, train all models with various \ac{at} methods, and evaluate on CIFAR (Sec. \ref{sec:cifar}) and ImageNet (Sec. \ref{sec:imagenet}) to verify the generalization of our design principles.
Full comparisons with \acp{cnn}, Transformers, and \ac{nas}-based architectures are provided in supplementary materials Sec.~\ref{sec:appx-cifar} and \ref{sec:appx-imagenet}.

\subsection{Combining the Three Design Principles}
\label{sec:roadmap}

We robustify a \ac{cnn} by applying our three architectural design principles.  
Specifically, we train ResNet-50 with Fast-\ac{at}~\cite{wong2020fast} and use its evaluation results on ImageNet as the starting point.
Table \ref{tab:roadmap} outlines the path we take to robustify ResNet-50, and total parameters are generally controlled over the course of exploration.
For the four body stages, we set $D_{1,2,3,4} = 5,8,13,1$ and $W_{1,2,3,4} = 36,72,140,270$ to control the \ac{wd} ratio within the optimal range. 
The convolutional stem stage incorporates the postponed downsampling operation and widens the output channels to 96.
Finally, we append the \ac{se} block ($r = 4$) to the $3 \times 3$ convolution layer and replace \ac{relu} with \ac{silu}. 
The clean and \ac{pgd} accuracies of \hlbox{\ac{ra}}ResNet-50 consistently improve for each modification, which verifies that our design principles work well both individually and collectively. 

\subsection{Evaluations on CIFAR-10 \& CIFAR-100}
\label{sec:cifar}

We apply all three principles to models within the design spaces of ResNet and \ac{wrn}. 
Detail specifications of each robustified architecture are in Sec.~\ref{sec:appx-arch} in supplementary materials. 
Table \ref{tab:results-cifar} presents the comprehensive evalution results on CIFAR-10 and CIFAR-100 against \ac{aa} and 20-step \ac{pgd} ($\ac{pgd}^{20}$) attacks with the same maximum perturbation $\ell_\infty, \epsilon = 8/255$. 
The training methods are \ac{sat}~\cite{madry2018towards}, TRADES~\cite{zhang2019theoretically}, MART~\cite{wang2020improving}, and diffusion-augmented \ac{at}~\cite{wang2023better}.
The diffusion-augmented \ac{at} is the \ac{sota} training recipe proposed by Wang \etal~\cite{wang2023better}, which augments the original CIFAR only with images generated by \ac{edm}~\cite{karras2022elucidating}, so that no external datasets are needed.
Since this training recipe incurs extreme computational costs~\cite{wang2023better}, we train on the 1M generated dataset using batch size 512 and epoch 400, abbreviated as ``Diff. 1M'' in Table \ref{tab:results-cifar}. 
In general, the gains are consistent across \ac{at} methods, parameter budgets, and design spaces on CIFAR-10 and CIFAR-100. 
Importantly, our design principles augment the improvements in network robustness achieved through better training schemes, boosting even the robustness of the \ac{sota} Diff. 1M method by 1--3 \ac{pp}.
Our best model, \hlbox{\ac{ra}}\ac{wrn}-70-16, achieves 66.33\% and 38.73\% \ac{aa} accuracy on CIFAR-10 and CIFAR-100, respectively. 
Due to limited computational resources, we only train ``Diff. 50M'' on CIFAR-10. 
Our \hlbox{\ac{ra}}\ac{wrn}-70-16 consistently outperforms \ac{wrn}-70-16 and achieves the best performance on RobustBench~\cite{croce2020robustbench}.

\subsection{Evaluations on ImageNet}
\label{sec:imagenet}

Similarly, we also apply all three principles to models within the design spaces of ResNet and \ac{wrn}. 
The training methods are Fast-\ac{at}~\cite{wong2020fast} and \ac{sat}~\cite{madry2018towards}.
Table \ref{tab:results-imagenet} presents the comprehensive \ac{sat} evaluation results on ImageNet against \ac{aa} and 100-step \ac{pgd} ($\ac{pgd}^{100}$). 
The maximum perturbation for \ac{aa} is $\ell_\infty, \epsilon = 4/255$, and for \ac{pgd} are $\ell_\infty, \epsilon = \{ 2,4,8 \}/255$. 
We observe a consistent 4--9 \ac{pp} robustness gain across different model parameters and design spaces. 
Specifically, our \hlbox{\ac{ra}}\ac{wrn}-101-2, even with fewer parameters than \ac{wrn}-101-2, improves the \ac{aa} accuracy by 6.94 \ac{pp} when trained from scratch with \ac{sat}.

\begin{table}[!htbp]
\renewcommand{\arraystretch}{0.3}
\small
\centering
\caption{Adversarial robustness on ImageNet against \ac{aa} and 100-step \ac{pgd} ($\ac{pgd}^{100}$). 
The maximum perturbation for \ac{aa} is $\ell_\infty, \epsilon = 4/255$, and for \ac{pgd} are $\ell_\infty, \epsilon = \{ 2,4,8 \}/255$. 
All models trained from random initializations with \ac{sat}~\cite{madry2018towards}. 
We observe a consistent 4--9 \ac{pp} robustness gain.
Full comparisons including Transformers and Fast-\ac{at} results are in supplementary materials Sec.~\ref{sec:appx-imagenet}.
}
\scalebox{0.80}{
\begin{tabular}{lr@{\hspace{6mm}}r@{\hspace{1mm}}lr@{\hspace{1mm}}lr@{\hspace{1mm}}lr@{\hspace{1mm}}lr@{\hspace{1mm}}l}
\toprule
    Model & \#Param. & \multicolumn{2}{c}{Clean (\%)} & \multicolumn{2}{c}{\ac{aa} (\%)} & \multicolumn{2}{c}{$\ac{pgd}^{100} \mh 2$ (\%)} & \multicolumn{2}{c}{$\ac{pgd}^{100} \mh 4$ (\%)} & \multicolumn{2}{c}{$\ac{pgd}^{100} \mh 8$ (\%)} \\
\midrule
    ResNet-50 & 26M & 63.87 & & 34.96 & & 52.15 & & 38.96 & & 15.83 \\
    \hlbox{\ac{ra}}ResNet-50 & 26M & \textbf{70.17} & \textcolor{mygreen}{+6.30} & \textbf{44.14} & \textcolor{mygreen}{+9.18} & \textbf{60.06} & \textcolor{mygreen}{+7.91} 
 & \textbf{47.77} & \textcolor{mygreen}{+8.81} & \textbf{21.77} & \textcolor{mygreen}{+5.94} \\
\midrule
    ResNet-101 & 45M & 67.06 & & 39.78 & & 56.26 & & 43.17 & & 18.31 \\
    \hlbox{\ac{ra}}ResNet-101 & 46M & \textbf{71.88} & \textcolor{mygreen}{+4.82} & \textbf{46.26} & \textcolor{mygreen}{+6.48} & \textbf{61.89} & \textcolor{mygreen}{+5.63} & \textbf{49.30} & \textcolor{mygreen}{+6.13} & \textbf{23.01} & \textcolor{mygreen}{+4.70} \\
\midrule
    \ac{wrn}-101-2 & 127M & 69.30 & & 42.00 & & 58.71 & & 45.27 & & 19.95 \\
    \hlbox{\ac{ra}}\ac{wrn}-101-2 & 104M & \textbf{73.44} & \textcolor{mygreen}{+4.14} & \textbf{48.94} & \textcolor{mygreen}{+6.94} & \textbf{63.49} & \textcolor{mygreen}{+4.78} & \textbf{51.03} & \textcolor{mygreen}{+5.76} & \textbf{25.31} & \textcolor{mygreen}{+5.36} \\
\bottomrule
\end{tabular}
}
\vspace{-5mm}
\label{tab:results-imagenet}
\end{table}
\section{Conclusion}
\label{sec:conclusion}

We synthesize a suite of three generalizable robust architectural design principles: 
(a) optimal range for \textit{depth} and \textit{width} configurations, 
(b) preferring \textit{convolutional} over \textit{patchify} stem stage, and 
(c) robust residual block design through adopting \acl{se} blocks and non-parametric smooth activation functions.
Through extensive experiments across a wide spectrum of \textit{dataset scales}, \textit{\acl{at} methods}, \textit{model parameters}, and \textit{network design spaces}, our principles consistently and markedly improve adversarial robustness on CIFAR-10, CIFAR-100, and ImageNet. 

\section{Acknowledgement}
\label{sec:aclnowledgement}

This work was supported in part by the Defense Advanced Research Projects Agency (DARPA). Use, duplication, or disclosure is subject to the restrictions as stated in Agreement number HR00112030001 between the Government and the Performer. This work was also supported in part by gifts from Avast, Fiddler Labs, Bosch, Facebook, Intel, NVIDIA, Google, Symantec, and Amazon.

\bibliography{egbib}

\newpage
\appendix
\section{Extended Description of Robust Architectures}
\label{sec:appx-related-work}

A few research studies have examined the impact of architectural designs on adversarial robustness~\cite{xie2019intriguing,guo2020meets,singh2023revisiting,huang2022revisiting}, \eg, RobustArt~\cite{tang2021robustart} is the first comprehensive robustness benchmark of architectures and training techniques on ImageNet variants and Jung \etal~\cite{jung2023neural} presented the first robustness dataset evaluating a complete \ac{nas} search space and demonstrated architectures’ impacts on robustness.
Among them, Huang \etal~\cite{huang2022revisiting} is the closest to ours, and we thus provide a more detailed comparison with their work. 
Similar to our work, Huang \etal~\cite{huang2022revisiting} also explored the relationship among depth and width, the \ac{se} block, and adversarial robustness through adversarially trained networks. However, our work is markedly different and enhanced in the following aspects:
\begin{enumerate}[leftmargin=*,topsep=0pt]
\itemsep-0.3em 
\item Huang \etal assigned a fixed depth and width ratio only for the 3-stage \ac{wrn} on CIFAR-10. 
It was an open research question as to how to extrapolate this fixed ratio to networks with more than three stages, such as the commonly used 4-stage residual networks for ImageNet~\cite{he2016deep,liu2022convnet,radosavovic2020designing}. 
In contrast, we provide a flexible compound scaling rule that does not place a restriction on the total number of stages, and we verify its generalizability and optimality through extensive experiments on CIFAR-10, CIFAR-100, and ImageNet.

\item Huang \etal proposed a specific residual block design using hierarchically aggregated convolution and residual \ac{se}. 
However, we show that such a residual \ac{se} is unnecessary due to the negative correlation between reduction ratio $r$ and robustness. 
Furthermore, our design principles are applicable to both basic and bottleneck residual block designs. This flexibility is advantageous since the basic block is commonly used on CIFAR and the bottleneck block is widely deployed on ImageNet to reduce computational complexity~\cite{he2016deep,he2016identity}.

\item Huang \etal found that the adversarial robustness of models with smooth activation functions was sensitive to \ac{at} hyperparameters, and that removing \ac{bn} affine parameters from weight decay was crucial; 
if the \ac{bn} affine parameters were not removed, smooth activation functions did not improve performance beyond that of \ac{relu}. 
This finding contradicts the prevailing consensus in the literature that smooth activation functions significantly improve robustness~\cite{xie2020smooth,bai2021transformers}. 
In our research, we also find that using smooth activation functions is beneficial to robustness, and removing the \ac{bn} affine parameters from weight decay is the correct implementation supported by Van~\cite{van2017l2} and multiple popular code bases and forums.\footnote{\href{https://github.com/karpathy/minGPT/pull/24\#issuecomment-679316025}{minGPT: A PyTorch re-implementation of GPT}}\textsuperscript{\textcolor{red}{,}}\footnote{\href{https://discuss.pytorch.org/t/weight-decay-in-the-optimizers-is-a-bad-idea-especially-with-batchnorm/16994}{PyTorch forum: Weight decay in the optimizers is a bad idea (especially with BatchNorm)}}

\item Finally, Huang \etal only explored on CIFAR-10 and CIFAR-100, leaving no evidence that these findings will extrapolate to the large-scale ImageNet. 
In contrast, we verified the generalization of all our design principles through extensive experiments over
a wide spectrum of dataset scales, \ac{at} methods, model parameters, and network design
spaces. 

\end{enumerate}

\smallskip
A parallel line of related studies leverages \ac{nas} to search for optimal robust architectures.
Guo \etal~\cite{guo2020meets} explored two types of block topologies within the DARTS search space. 
Subsequent work regulated the \ac{nas} loss formulation through the smoothness of the input loss landscape~\cite{mok2021advrush}. 
These compute-intensive \ac{nas} frameworks mainly focus on searching for block topology while leaving other factors to manual design, \eg, activation, depth, and width. 
Furthermore, most searches are conducted on CIFAR-10 since both \ac{nas} and \ac{at} are already computationally expensive. 
Besides, current research shows that the 
\ac{nas}-optimized architecture depends on the dataset used~\cite{lee2021rapid},
thus hindering explorations of network design principles that deepen our understanding and generalize to new settings~\cite{radosavovic2020designing}.  
In Sec.~\ref{sec:appx-cifar}, we also compare our results with \ac{nas}-based networks and demonstrate that our robustified networks exhibit higher robustness.

\section{Additional Details on Architectural Design Principles}
\label{sec:appx-principle}

\subsection{Full Results for Stem Stage and Residual Block Designs}
\label{sec:appx-other-components}

Here we provide the full results in Table~\ref{tab:appx-components} showcasing how various configurations of \textit{stem stage} and \textit{residual block designs} impact the clean and adversarial accuracies over the ResNet-50 baseline, extending the result highlights presented in Figure~\ref{fig:other-components} of the main paper.

For the stem stage, the convolution stem with postponed downsampling operation outperforms the patchify stem. 
In the patchify stem, we observe a consistent performance improvement while decreasing the stride, and the $4 \times 4$ patch with the largest overlapping between neighboring patches (Patch 4, Stride 1) performs almost on par with the postponed downsampling.
Finally, decreasing the width from 64 (ResNet-50) to 32 lowers the accuracy due to fewer model parameters, while increasing the width from 64 to 96 significantly boosts both clean and adversarial accuracies with a negligible 0.01M increase in total parameters. 

For the residual block design, we find that a straightforward application of \ac{se} markedly increases all \ac{pgd} and clean accuracies compared to ResNet-50. 
In terms of activation, reducing the number of activation layers does not contribute to adversarial robustness. 
Therefore, we preserve the activations along with all convolutions. 
Besides, we find that non-parametric smooth activation functions exhibit greater robustness compared to both \ac{relu} and their parametric counterparts. 

\begin{table}[!htbp]
\renewcommand{\arraystretch}{0.9}
\small
\centering
\caption{Full results showcasing how various configurations of \textit{stem stage} and \textit{residual block designs} impact the clean and adversarial accuracies over the ResNet-50 baseline, extending the result highlights already presented in Figure~\ref{fig:other-components} of the main paper. All models trained with Fast-AT~\cite{wong2020fast} and evaluated on full ImageNet validation set.}
\scalebox{0.80}{
\begin{tabular}{cl@{\hspace{6mm}}r@{\hspace{1mm}}l@{\hspace{6mm}}r@{\hspace{1mm}}l@{\hspace{6mm}}r@{\hspace{1mm}}l@{\hspace{6mm}}r@{\hspace{1mm}}l}
\toprule
    & Config & \multicolumn{2}{c}{Clean (\%)} & \multicolumn{2}{c}{$\acs{pgd}^{10} \mh 2$ (\%)} & \multicolumn{2}{c}{$\acs{pgd}^{10} \mh 4$ (\%)} & \multicolumn{2}{c}{$\acs{pgd}^{10} \mh 8$ (\%)} \\
\midrule
    & ResNet-50 (baseline) & 56.05 & & 42.81 & & 30.59 & & 12.62 \\
\midrule
\multirow{8}{*}{Stem stage} & Postponed downsampling & 57.08 & \textcolor{mygreen}{+1.03}    & 45.19 & \textcolor{mygreen}{+2.38} & 33.08 & \textcolor{mygreen}{+2.49} & 14.50 & \textcolor{mygreen}{+1.88} \\
& Patch 4, Stride 4      & 55.40 & \textcolor{mypurple}{\ -0.65} & 43.45 & \textcolor{mygreen}{+0.64} & 31.68 & \textcolor{mygreen}{+1.09} & 13.80 & \textcolor{mygreen}{+1.18} \\
& Patch 2, Stride 2      & 56.38 & \textcolor{mygreen}{+0.33}    & 44.21 & \textcolor{mygreen}{+1.40} & 31.91 & \textcolor{mygreen}{+1.32} & 13.48 & \textcolor{mygreen}{+0.86} \\
& Patch 4, Stride 3      & 55.75 & \textcolor{mypurple}{\ -0.30} & 44.54 & \textcolor{mygreen}{+1.73} & 32.41 & \textcolor{mygreen}{+1.82} & 13.74 & \textcolor{mygreen}{+1.12} \\
& Patch 4, Stride 2      & 56.58 & \textcolor{mygreen}{+0.53}    & 44.60 & \textcolor{mygreen}{+1.79} & 32.83 & \textcolor{mygreen}{+2.24} & 14.03 & \textcolor{mygreen}{+1.41} \\
& Patch 4, Stride 1      & 56.72 & \textcolor{mygreen}{+0.67}    & 45.06 & \textcolor{mygreen}{+2.25} & 33.20 & \textcolor{mygreen}{+2.61} & 14.45 & \textcolor{mygreen}{+1.83} \\
& Stem width $= 32$      & 55.89 & \textcolor{mypurple}{\ -0.16}         & 41.64 & \textcolor{mypurple}{\ -1.17} & 29.73 & \textcolor{mypurple}{\ -0.86}      & 13.25 & \textcolor{mygreen}{+0.63} \\
& Stem width $= 96$      & 57.29 & \textcolor{mygreen}{+1.24}         & 44.55 & \textcolor{mygreen}{+1.47} & 32.06 & \textcolor{mygreen}{+1.47}      & 13.74 & \textcolor{mygreen}{+1.12} \\
\midrule
\multirow{12}{*}{Residual block design} & \acs{se} ($r = 4$)                & 57.83 & \textcolor{mygreen}{+1.78} & 45.09 & \textcolor{mygreen}{+2.28} & 32.64 & \textcolor{mygreen}{+2.05} & 14.01 & \textcolor{mygreen}{+1.39} \\
& \acs{relu}$\mh$\acs{relu}$\mh$0   & 51.54 &  \textcolor{mypurple}{\ -4.51} & 38.69 &  \textcolor{mypurple}{\ -4.12} & 27.05 &  \textcolor{mypurple}{\ -3.54} & 10.94 &  \textcolor{mypurple}{\ -1.68} \\
& \acs{relu}$\mh$0$\mh$\acs{relu}   & 53.91 &  \textcolor{mypurple}{\ -2.14} & 41.22 &  \textcolor{mypurple}{\ -1.59} & 29.62 &  \textcolor{mypurple}{\ -0.97} & 12.30 &  \textcolor{mypurple}{\ -0.32} \\
& 0$\mh$\acs{relu}$\mh$\acs{relu}   & 54.81 &  \textcolor{mypurple}{\ -1.24} & 42.10 &  \textcolor{mypurple}{\ -0.71} & 30.34 &  \textcolor{mypurple}{\ -0.25} & 12.86 &  \textcolor{mygreen}{+0.24} \\
& 0$\mh$0$\mh$\acs{relu}            & 51.03 &  \textcolor{mypurple}{\ -5.02} & 39.12 &  \textcolor{mypurple}{\ -3.69} & 28.15 &  \textcolor{mypurple}{\ -2.44} & 12.09 &  \textcolor{mypurple}{\ -0.53} \\
& 0$\mh$\acs{relu}$\mh$0            & 47.18 &  \textcolor{mypurple}{\ -8.87} & 34.85 &  \textcolor{mypurple}{\ -7.96} & 24.12 &  \textcolor{mypurple}{\ -6.47} &  9.51 &  \textcolor{mypurple}{\ -3.11} \\
& \acs{relu}$\mh$0$\mh$0            & 44.21 &  \textcolor{mypurple}{-11.84} & 32.34 &  \textcolor{mypurple}{-10.47} & 22.24 &  \textcolor{mypurple}{\ -8.35} &  8.77 &  \textcolor{mypurple}{\ -3.85} \\
& \acs{gelu}                        & 57.48 & \textcolor{mygreen}{+1.43}    & 45.05 &  \textcolor{mygreen}{+2.24} & 33.12 & \textcolor{mygreen}{+2.53}    & 14.80 &  \textcolor{mygreen}{+2.18} \\
& \acs{silu}                        & 58.19 & \textcolor{mygreen}{+2.14}    & 46.21 &  \textcolor{mygreen}{+3.40} & 34.07 & \textcolor{mygreen}{+3.48}    & 14.68 &  \textcolor{mygreen}{+2.06} \\
& \acs{prelu}                       & 55.81 & \textcolor{mypurple}{\ -0.27} & 42.52 &  \textcolor{mypurple}{\ -0.29} & 30.38 & \textcolor{mypurple}{\ -0.21} & 12.76 &  \textcolor{mygreen}{+0.14} \\
& \acs{psilu}                       & 56.38 & \textcolor{mygreen}{+0.33}    & 44.90 &  \textcolor{mygreen}{+2.09} & 33.76 & \textcolor{mygreen}{+3.17}    & 15.40 &  \textcolor{mygreen}{+2.78} \\
& \acs{pssilu}                      & 57.43 & \textcolor{mygreen}{+1.38}    & 44.44 &  \textcolor{mygreen}{+1.63} & 32.22 & \textcolor{mygreen}{+1.63}    & 13.71 &  \textcolor{mygreen}{+1.09} \\
\bottomrule
\end{tabular}
}
\label{tab:appx-components}
\end{table}

\newpage
\subsection{Adversarial Robustness Negatively Correlated with SE Reduction Ratio \textit{r} on CIFAR-10}
\label{sec:appx-se}

\begin{table}[!htbp]
\renewcommand{\arraystretch}{0.9}
\small
\centering
\caption{
A hyperparameter sweep of the \ac{se} block reduction ratio $r = \{2, 4, 8, 16, 32, 64\}$ on CIFAR-10. 
These results show that adversarial robustness is negatively correlated with $r$, and when $r \geq 32$, the accuracy is inferior to the baseline WRN-22-10, supporting our discovery presented in Sec.~\ref{sec:se}. 
Both \ac{aa} and 20-step \ac{pgd} ($\ac{pgd}^{20}$) use the same maximum perturbation $\ell_\infty, \epsilon = 8/255$. 
}
\scalebox{0.82}{
\begin{tabular}{l@{\hspace{8mm}}r@{\hspace{1mm}}l@{\hspace{8mm}}r@{\hspace{1mm}}l@{\hspace{8mm}}r@{\hspace{1mm}}l}
\toprule
    Config & \multicolumn{2}{c}{Clean (\%)} & \multicolumn{2}{c}{\acs{aa} (\%)} & \multicolumn{2}{c}{$\acs{pgd}^{20}$ (\%)} \\
\midrule
    \acs{wrn}-22-10            & 83.82 & & 48.38 & & 52.64 \\
\midrule
    \acs{wrn}-22-10 ($r = 2$)  & 86.95 & \textcolor{mygreen}{+3.13} & 49.14 & \textcolor{mygreen}{+0.76} & 53.48 & \textcolor{mygreen}{+0.84} \\
    \acs{wrn}-22-10 ($r = 4$)  & 86.75 & \textcolor{mygreen}{+2.93} & 49.13 & \textcolor{mygreen}{+0.75} & 53.52 & \textcolor{mygreen}{+0.88} \\
    \acs{wrn}-22-10 ($r = 8$)  & 86.48 & \textcolor{mygreen}{+2.66} & 49.11 & \textcolor{mygreen}{+0.73} & 53.39 & \textcolor{mygreen}{+0.75} \\
    \acs{wrn}-22-10 ($r = 16$) & 84.97 & \textcolor{mygreen}{+1.15} & 48.89 & \textcolor{mygreen}{+0.51} & 53.04 & \textcolor{mygreen}{+0.40} \\
    \acs{wrn}-22-10 ($r = 32$) & 83.61 & \textcolor{mypurple}{\ -0.21} & 48.20 & \textcolor{mypurple}{\ -0.18} & 52.24 & \textcolor{mypurple}{\ -0.40} \\
    \acs{wrn}-22-10 ($r = 64$) & 82.90 & \textcolor{mypurple}{\ -0.92} & 47.44 & \textcolor{mypurple}{\ -0.94} & 51.43 & \textcolor{mypurple}{\ -1.21} \\
\bottomrule
\end{tabular}
}
\label{tab:appx-se}
\end{table}

\section{Evaluations on CIFAR-10 \& CIFAR-100}
\label{sec:appx-cifar}

This section presents the complete results of our robustified architectures in Table~\ref{tab:appx-cifar} and provides a systematic comparison to \ac{sota} adversarially trained Transformers and \ac{nas}-based networks in Table~\ref{tab:appx-cifar-sys},
extending the result highlights presented in Sec.~\ref{sec:cifar} in the main paper. 
As the \ac{at} recipes for \acp{cnn} and Transformers are not compatible with each other, we do not retrain the Transformers and instead, directly extract the results from the literature. 
In addition, the \ac{at} recipe for Transformers requires multiple training tricks built on \ac{sat} to boost robustness, \eg, attention random dropping~\cite{mo2022adversarial}, perturbation random masking~\cite{mo2022adversarial}, $\epsilon$-warmup~\cite{debenedetti2022light}, and larger weight decay~\cite{debenedetti2022light}. 
Despite employing all these tricks in training Transformers, our \hlbox{\ac{ra}}\ac{wrn}-34-12 trained with ``Diff. 1M'' is significantly more robust than Swin-B and \ac{xcit}-L12 on both CIFAR-10 and CIFAR-100 even with fewer total parameters. 
We also compare to \ac{nas}-based networks: our \hlbox{\ac{ra}}\ac{wrn}-22-10 is significantly more robust than RobNet-large-v2~\cite{guo2020meets} and performs on par with AdvRush~\cite{mok2021advrush} using the same TRADES~\cite{zhang2019theoretically} \ac{at} method but with fewer total parameters.
Lastly, we compare our robust architectures with Huang \etal~\cite{huang2021exploring}, who have also studied the relationship between robustness and depth and width, and proposed a reconfigured version of \ac{wrn}-34-12 called \ac{wrn}-34-R. 
By using the same \ac{sat} methods, both \hlbox{\ac{ra}}\ac{wrn}-34-12 and \ac{wrn}-34-R show greater robustness than the baseline \ac{wrn}-34-12, but our \hlbox{\ac{ra}}\ac{wrn}-34-12 is 1.75 \ac{pp} and 0.69 \ac{pp} higher than \ac{wrn}-34-R in terms of \ac{aa} and \ac{pgd} accuracies, respectively.

\begin{table}[!htbp]
\renewcommand{\arraystretch}{0.85}
\small
\centering
\caption{Complete results of adversarial robustness on CIFAR-10 and CIFAR-100 against \ac{aa} and 20-step \ac{pgd} ($\ac{pgd}^{20}$) with the same maximum perturbation $\ell_\infty, \epsilon = 8/255$. 
Applying our principles leads to a consistent 1--3 \ac{pp} robustness gain across \ac{at} methods, parameter budgets, and design spaces,
boosting even the \ac{sota} ``Diff. 1M'' and ``Diff. 50M'' \ac{at} methods proposed by Wang \etal~\cite{wang2023better}. 
This table extends Table~\ref{tab:results-cifar} in the main paper by including the results for \ac{wrn}-22-10.
}
\scalebox{0.76}{
\begin{tabular}{@{}l@{\hspace{2mm}}l@{\hspace{2mm}}l|r@{\hspace{1mm}}l@{\hspace{2mm}}r@{\hspace{1mm}}l@{\hspace{2mm}}r@{\hspace{1mm}}l|r@{\hspace{1mm}}l@{\hspace{2mm}}r@{\hspace{1mm}}l@{\hspace{2mm}}r@{\hspace{1mm}}l@{}}
\toprule
    \multirow{2}{*}{\#Param.} & \multirow{2}{*}{Method} & \multirow{2}{*}{Model} & \multicolumn{6}{c}{CIFAR-10} & \multicolumn{6}{c}{CIFAR-100}\\
   & & & \multicolumn{2}{c}{Clean (\%)} & \multicolumn{2}{c}{\acs{aa} (\%)} & \multicolumn{2}{c}{$\ac{pgd}^{20}$ (\%)} & \multicolumn{2}{c}{Clean (\%)} & \multicolumn{2}{c}{\acs{aa} (\%)} & \multicolumn{2}{c}{$\ac{pgd}^{20}$ (\%)} \\
\midrule
    \multirow{8}{*}{26M} & \multirow{2.5}{*}{SAT} & ResNet-50 & 84.05 & & 49.97 & & 54.37 & & 55.86 & & 23.78 & & 27.48 \\
    & & \hlbox{\acs{ra}}ResNet-50 & \textbf{84.91} & \textcolor{mygreen}{+0.86} & \textbf{50.94} & \textcolor{mygreen}{+0.97} & \textbf{55.19} & \textcolor{mygreen}{+0.82} & \textbf{56.38} & \textcolor{mygreen}{+0.52} & \textbf{24.99} & \textcolor{mygreen}{+1.21} & \textbf{28.84} & \textcolor{mygreen}{+1.36} \\
\cmidrule{2-15}
    & \multirow{2.5}{*}{TRADES} & ResNet-50 & 82.26 & & 49.91 & & 54.50 & & 56.00 & & 25.05 & & 29.91 \\
    & & \hlbox{\acs{ra}}ResNet-50 & \textbf{82.80} & \textcolor{mygreen}{+0.54} & \textbf{51.23} & \textcolor{mygreen}{+1.32} & \textbf{55.44} & \textcolor{mygreen}{+0.94} & \textbf{56.29} & \textcolor{mygreen}{+0.29} & \textbf{25.83} & \textcolor{mygreen}{+0.78} & \textbf{31.87} & \textcolor{mygreen}{+1.96} \\
\cmidrule{2-15}
    & \multirow{2.5}{*}{MART} & ResNet-50 & 77.98 & & 47.17 & & 52.70 & & 53.18 & & 25.35 & & 30.79 \\
    & & \hlbox{\acs{ra}}ResNet-50 & \textbf{79.60} & \textcolor{mygreen}{+1.62} & \textbf{49.19} & \textcolor{mygreen}{+2.02} & \textbf{56.47} & \textcolor{mygreen}{+3.77} & \textbf{53.68} & \textcolor{mygreen}{+0.50} & \textbf{26.97} & \textcolor{mygreen}{+1.62} & \textbf{32.81} & \textcolor{mygreen}{+2.02} \\
\midrule
    \multirow{5}{*}{27M} & \multirow{2.5}{*}{SAT} & \ac{wrn}-22-10 & 83.82 & & 48.38 & & 52.64 & & 56.79 & & 23.46 & & 27.08 \\
    & & \hlbox{\ac{ra}}\ac{wrn}-22-10 & \textbf{84.27} & \textcolor{mygreen}{+0.45} & \textbf{51.30} & \textcolor{mygreen}{+2.92} & \textbf{55.42} & \textcolor{mygreen}{+2.78} & \textbf{57.34} & \textcolor{mygreen}{+0.55} & \textbf{24.27} & \textcolor{mygreen}{+0.81} & \textbf{28.64} & \textcolor{mygreen}{+1.56} \\
\cmidrule{2-15}
    & \multirow{2.5}{*}{TRADES} & \ac{wrn}-22-10 & 81.81 & & 51.06 & & 55.21 & & 55.48 & & 23.50 & & 29.60 \\
    & & \hlbox{\ac{ra}}\ac{wrn}-22-10 & \textbf{82.27} & \textcolor{mygreen}{+0.46} & \textbf{51.71} & \textcolor{mygreen}{+0.65} & \textbf{56.20} & \textcolor{mygreen}{+0.99} & \textbf{55.55} & \textcolor{mygreen}{+0.07} & \textbf{24.91} & \textcolor{mygreen}{+1.41} & \textbf{29.78} & \textcolor{mygreen}{+0.18} \\
\midrule
    \multirow{11}{*}{37M} & \multirow{2.5}{*}{SAT} & \ac{wrn}-28-10 & 85.44 & & 48.45 & &  53.13 & & \textbf{60.49} & & 23.64 & & 27.47 \\
    & & \hlbox{\ac{ra}}\ac{wrn}-28-10 & \textbf{85.52} & \textcolor{mygreen}{+0.08} & \textbf{51.96} & \textcolor{mygreen}{+3.51} & \textbf{56.22} & \textcolor{mygreen}{+3.09} & 59.09 & \textcolor{mypurple}{-1.40} & \textbf{25.14} & \textcolor{mygreen}{+1.50} & \textbf{29.27} & \textcolor{mygreen}{+1.80}  \\
\cmidrule{2-15}
    & \multirow{2.5}{*}{TRADES} & \ac{wrn}-28-10 & \textbf{83.86} & & 51.79 & & 55.69 & & 55.21 & & 25.47 & & 29.34 \\
    & & \hlbox{\ac{ra}}\ac{wrn}-28-10 & 83.29 & \textcolor{mypurple}{-0.57} & \textbf{52.10} & \textcolor{mygreen}{+0.31} & \textbf{56.31} & \textcolor{mygreen}{+0.62} & \textbf{55.38} & \textcolor{mygreen}{+0.71} & \textbf{25.68} & \textcolor{mygreen}{+0.21} & \textbf{29.41} & \textcolor{mygreen}{+0.07} \\
\cmidrule{2-15}
    & \multirow{2.5}{*}{MART} & \ac{wrn}-28-10 & 82.83 & & 50.30 & & 57.00 & & 51.31 & & 25.78 & &  30.06\\
    & & \hlbox{\ac{ra}}\ac{wrn}-28-10 & \textbf{82.85} & \textcolor{mygreen}{+0.02} & \textbf{50.81} & \textcolor{mygreen}{+0.51} & \textbf{57.35} & \textcolor{mygreen}{+0.35} & \textbf{51.61} & \textcolor{mygreen}{+0.30} & \textbf{26.11} & \textcolor{mygreen}{+0.33} & \textbf{30.82} & \textcolor{mygreen}{+0.76} \\
\cmidrule{2-15}
    & \multirow{2.5}{*}{Diff. 1M} & \ac{wrn}-28-10 & 90.61 & & 61.66 & & 66.43 & & 67.26 & & 34.26 & & 39.29 \\
    & & \hlbox{\ac{ra}}\ac{wrn}-28-10 & \textbf{91.32} & \textcolor{mygreen}{+0.71} & \textbf{65.11} & \textcolor{mygreen}{+3.45} & \textbf{68.93} & \textcolor{mygreen}{+2.50} & \textbf{69.03} & \textcolor{mygreen}{+1.77} & \textbf{37.24} & \textcolor{mygreen}{+2.98} & \textbf{41.59} & \textcolor{mygreen}{+2.30} \\
\midrule
    \multirow{5}{*}{67M} & \multirow{2.5}{*}{SAT} & \ac{wrn}-34-12 & 85.92 & & 49.35 & & 53.05 & & 59.08 & & 23.69 & & 27.05   \\
    & & \hlbox{\ac{ra}}\ac{wrn}-34-12 & \textbf{86.50} & \textcolor{mygreen}{+0.58} & \textbf{51.78} & \textcolor{mygreen}{+2.43} & \textbf{56.04} & \textcolor{mygreen}{+2.99} & \textbf{59.46} & \textcolor{mygreen}{+0.38} & \textbf{25.18} & \textcolor{mygreen}{+1.49} & \textbf{29.49} & \textcolor{mygreen}{+2.44} \\
\cmidrule{2-15}
    & \multirow{2.5}{*}{Diff. 1M} & \ac{wrn}-34-12 & 91.11 & & 62.83 & & 67.53 & & 68.40 & & 35.67 & & 40.33 \\
    & & \hlbox{\ac{ra}}\ac{wrn}-34-12 & \textbf{91.75} & \textcolor{mygreen}{+0.64} & \textbf{65.71} & \textcolor{mygreen}{+2.88} & \textbf{69.67} & \textcolor{mygreen}{+2.14} & \textbf{69.75} & \textcolor{mygreen}{+1.35} & \textbf{37.73} & \textcolor{mygreen}{+2.06} & \textbf{42.16} & \textcolor{mygreen}{+1.83} \\
\midrule
    \multirow{8.5}{*}{267M} & \multirow{2.5}{*}{SAT} & \ac{wrn}-70-16 & 86.26 & & 50.19 & & 53.74 & & 60.26 & & 23.99 & & 27.05\\
    & & \hlbox{\acs{ra}}\ac{wrn}-70-16  & \textbf{86.72} & \textcolor{mygreen}{+0.46} & \textbf{52.13} & \textcolor{mygreen}{+1.94} & \textbf{56.49} & \textcolor{mygreen}{+2.75} & \textbf{60.42} & \textcolor{mygreen}{+0.16} & \textbf{25.17} & \textcolor{mygreen}{+1.18} & \textbf{29.46} & \textcolor{mygreen}{+2.41}\\
\cmidrule{2-15}
    & \multirow{2.5}{*}{Diff. 1M} & \ac{wrn}-70-16 & 91.82 & & 65.02 & & 69.10 & & 70.10 & & 37.77  & & 41.95\\
    & & \hlbox{\acs{ra}}\ac{wrn}-70-16 & \textbf{92.16} & \textcolor{mygreen}{+0.34} & \textbf{66.33} & \textcolor{mygreen}{+1.31} & \textbf{70.37} & \textcolor{mygreen}{+1.27} & \textbf{70.25} & \textcolor{mygreen}{+0.15} & \textbf{38.73} & \textcolor{mygreen}{+0.96} & \textbf{42.61} & \textcolor{mygreen}{+0.66} \\
\cmidrule{2-15}
    & \multirow{2.5}{*}{Diff. 50M} & \ac{wrn}-70-16 & 93.25 & & 70.69 & & 73.89 & & - & & -  & & - \\
    & & \hlbox{\acs{ra}}\ac{wrn}-70-16 & \textbf{93.27} & \textcolor{mygreen}{+0.02} & \textbf{71.07} & \textcolor{mygreen}{+0.38} & \textbf{75.28} & \textcolor{mygreen}{+1.39} & - &  & - & & - & \\
\bottomrule
\end{tabular}
}
\label{tab:appx-cifar}
\end{table}
\begin{table}[!htbp]
\renewcommand{\arraystretch}{0.9}
\small
\centering
\caption{
A systematic comparison to \ac{sota} adversarially trained Transformers and \ac{nas}-based architectures with adversarial robustness on CIFAR-10 and CIFAR-100 against \ac{aa} and 20-step \ac{pgd} ($\ac{pgd}^{20}$) with the same maximum perturbation $\ell_\infty, \epsilon = 8/255$. 
Our robustified architectures (prefixed by \hlbox{\acs{ra}}) exhibit greater robustness (highlighted in \textbf{bold}) than all Transformers and \ac{nas}-based architectures compared.  
The ``Diff. 1M'' results are extracted from Table~\ref{tab:appx-cifar}.
}
\scalebox{0.76}{
\begin{tabular}{@{}l@{\hspace{1mm}}l@{\hspace{1mm}}r|rrr|rrr@{}}
\toprule
    \multirow{2}{*}{Method} & \multirow{2}{*}{Model} & \multirow{2}{*}{\#Param.} & \multicolumn{3}{c}{CIFAR-10} & \multicolumn{3}{c}{CIFAR-100}\\
   & & & Clean (\%) & \acs{aa} (\%) & $\ac{pgd}^{20}$ (\%) & Clean (\%) & \acs{aa} (\%) & $\ac{pgd}^{20}$ (\%) \\
\midrule
    \multirow{3.5}{*}{Diff. 1M} & \hlbox{\acs{ra}}\acs{wrn}-28-10 &  37M & 91.32 & \textbf{65.11} & \textbf{68.93} & 69.03 & \textbf{37.24} & \textbf{41.59} \\
     & \hlbox{\acs{ra}}\acs{wrn}-34-12 &  67M & 91.75 & \textbf{65.71} & \textbf{69.67} & 69.75 & \textbf{37.73} & \textbf{42.16} \\
     & \hlbox{\acs{ra}}\acs{wrn}-70-16 & 267M & 92.16 & \textbf{66.33} & \textbf{70.37} & 70.25 & \textbf{38.73} & \textbf{42.61} \\
\midrule
    \multirow{4}{*}{Mo \etal~\cite{mo2022adversarial}} & \acs{deit}-S & 22M & 83.04 & 48.34 & 52.52 & - & - & -  \\
     & Swin-S        & 50M & 84.46 & 46.17 & 50.02 & - & - & - \\
     & \ac{vit}-B/16 & 86M & 84.90 & 50.03 & 53.80 & - & - & - \\
     & Swin-B        & 88M & 84.16 & 47.50 & 51.47 & - & - & - \\
\midrule
    \multirow{3}{*}{Debenedetti \etal~\cite{debenedetti2022light}} & \acs{xcit}-S12 & 26M & 90.06 & 56.14 & - & 67.34 & 32.19 & - \\
    & \acs{xcit}-M12 & 46M & 91.30 & 57.27 & - & 69.21 & 34.21 & -  \\ 
    & \acs{xcit}-L12 & 104M & 91.73 & 57.58 & - & 70.76 & 35.08 & - \\ 
\midrule
    TRADES & RobNet-large-v2~\cite{guo2020meets}   & 33M & 84.57 & 47.48 & 52.79 & 55.27 & 23.69 & 29.23 \\
    TRADES & AdvRush (7@96)~\cite{mok2021advrush}  & 33M & 84.65 & 52.08 & 56.23 & 55.40 & 25.27 & 29.40 \\
    SAT & \acs{wrn}-34-R~\cite{huang2021exploring} & 68M & 87.85 & 50.03 & 55.35 & 61.33 & 25.20 & 29.02 \\
\bottomrule
\end{tabular}
}
\label{tab:appx-cifar-sys}
\end{table}

\newpage
\section{Evaluations on ImageNet}
\label{sec:appx-imagenet}

This section presents an extended discussion of the ImageNet results in Sec.~\ref{sec:imagenet} in the main paper. 
Table~\ref{tab:appx-fat} provides a controlled comparison of our robustified architectures to \ac{sota} \acp{cnn} and Transformers using Fast-\ac{at} method~\cite{wong2020fast}. 
We present the robustified ResNet-50, ResNet-101, and \ac{wrn}-101-2 and make the following observations:
\begin{enumerate}[leftmargin=*,topsep=0pt]
\itemsep-0.3em 
\item Our robustified architectures consistently demonstrate a 4--9 \ac{pp} gain in robustness across different model parameters and design spaces. 
Furthermore, increasing the total number of parameters in general leads to higher robustness.
\item Under a fixed model capacity, our \hlbox{\ac{ra}}ResNet-50 outperforms the baseline ResNet-50 and ResNeXt-50 32$\times$4d~\cite{xie2017aggregated}, and \hlbox{\ac{ra}}ResNet-101 outperforms ResNet-101 and RegNetX-8GF~\cite{radosavovic2020designing}. 
\item Compared to models with larger parameters, our \hlbox{\ac{ra}}ResNet-50 is more robust than ResNet-152 and \ac{wrn}-50-2, and even \ac{wrn}-101-2 despite having $4.85\times$ fewer parameters. Similarly, our \hlbox{\ac{ra}}\ac{wrn}-101-2 outperforms the baseline \ac{wrn}-101-2 and achieves \ac{sota} performance under the Fast-\ac{at} method. 
\item Transformers such as Swin-T~\cite{liu2021swin} and Transformer-based architectures such as ConvNeXt-T~\cite{liu2022convnet} exhibit lower robustness when employing Fast-\ac{at}. The phenomenon can be attributed to the differences in optimizers, learning rates, and data augmentation, where most Transformer-related architectures use AdamW~\cite{loshchilov2017decoupled}, tiny learning rates, and heavy data augmentation.
\end{enumerate}

\smallskip
Then, we provide a systematic comparison of our \ac{sat}-trained robust architectures with \acp{cnn} and Transformers that utilize specifically optimized \ac{at} methods in Table~\ref{tab:appx-imagenet}. 
By applying our design principles, the robustified architecture achieves a similar or even superior level of robustness compared to the Transformers that utilize additional training tricks to enhance their robustness.
For example, under similar total parameters, ResNet-50 and ResNet-101 are less robust than \ac{xcit}-S12 and \ac{xcit}-M12, respectively, but the robustified \hlbox{\ac{ra}}ResNet-50 and \hlbox{\ac{ra}}ResNet-101 show higher clean and \ac{aa} accuracies. 
Additionally, there is no sign of saturation when scaling up the total parameters, as \hlbox{\ac{ra}}\ac{wrn}-101-2 remains markedly more robust than \ac{xcit}-L12 with the same 104M parameters. 

\begin{table}
\small
\centering
\caption{
Our robustified architectures (\hlbox{\ac{ra}}) consistently demonstrate a 4--9 \ac{pp} gain in robustness over \ac{sota} \acp{cnn} and Transformers using Fast-\ac{at} method~\cite{wong2020fast}, across different model parameters, design spaces, and attack budgets. 
}
\scalebox{0.85}{
\begin{tabular}{lrrrrr}
\toprule
     Model & \#Param. & Clean (\%) & $\ac{pgd}^{10} \mh 2$ (\%) & $\ac{pgd}^{10} \mh 4$ (\%) & $\ac{pgd}^{10} \mh 8$ (\%) \\
\midrule
    \hlbox{\ac{ra}}ResNet-50       & 26M & 62.02 & 51.47 & 39.65 & 18.97 \\
    \hlbox{\ac{ra}}ResNet-101      & 46M & 64.40 & 53.97 & 42.06 & 20.98 \\
    \hlbox{\ac{ra}}\ac{wrn}-101-2 & 104M & 66.08 & 55.52 & 43.81 & 22.50 \\

\midrule
    SqueezeNet 1.1            &  1M & 0.10  & 0.10  & 0.10  & 0.10  \\
    MobileNet V2              &  4M & 41.60 & 31.23 & 21.89 & 8.94  \\
    EfficientNet-B0           &  5M & 48.78 & 37.74 & 26.90 & 10.92 \\
    ShuffleNet V2 2.0$\times$ &  7M & 49.99 & 0.01  & 0.01  & 0.02  \\
    DenseNet-121              &  8M & 52.29 & 40.06 & 28.72 & 12.23 \\
    ResNet-18                 & 12M & 46.59 & 35.05 & 24.64 & 9.95  \\
    RegNetX-3.2GF             & 15M & 57.26 & 45.74 & 33.85 & 15.37 \\
    RegNetY-3.2GF             & 19M & 59.15 & 47.09 & 34.82 & 15.51 \\
    EfficientNetV2-S          & 21M & 57.64 & 45.89 & 33.48 & 14.03 \\
    ResNeXt-50 32$\times$4d   & 25M & 57.33 & 45.46 & 33.08 & 14.45 \\
    ResNet-50                 & 26M & 56.09 & 42.66 & 30.43 & 12.61 \\
    Swin-T                    & 28M & 38.83 & 28.08 & 18.49 & 6.20  \\
    ConvNeXt-T                & 29M & 21.35 & 15.39 & 10.51 & 4.07  \\
    DenseNet-161              & 29M & 59.80 & 47.60 & 35.35 & 15.77 \\
    EfficientNet-B5           & 30M & 55.90 & 44.80 & 33.26 & 14.53 \\
    RegNetY-8GF               & 39M & 63.61 & 52.26 & 40.15 & 19.21 \\
    RegNetX-8GF               & 40M & 60.26 & 48.98 & 36.89 & 17.22 \\
    ResNet-101                & 45M & 58.04 & 45.72 & 33.90 & 15.93 \\
    ResNet-152                & 60M & 61.55 & 48.50 & 35.85 & 15.87 \\
    \ac{wrn}-50-2             & 69M & 60.66 & 46.99 & 34.10 & 15.37 \\
    \ac{wrn}-101-2           & 127M & 61.63 & 49.10 & 36.23 & 16.14 \\
\bottomrule
\end{tabular}
}
\label{tab:appx-fat}
\end{table}
\begin{table}[!htbp]
\small
\centering
\caption{
By applying our design principles, the robustified architecture achieves a similar or even superior level of robustness compared to the Transformers that utilize additional training tricks to enhance their robustness.
The \hlbox{\ac{ra}} results are extracted from Table~\ref{tab:results-imagenet} in the main paper. 
}
\scalebox{0.85}{
\begin{tabular}
{lrrrrrr}
\toprule
    Model & \#Param. & Clean (\%) & \ac{aa} (\%) & $\ac{pgd}^{100} \mh 2$ (\%) & $\ac{pgd}^{100} \mh 4$ (\%) & $\ac{pgd}^{100} \mh 8$ (\%) \\
\midrule
    \hlbox{\ac{ra}}ResNet-50      & 26M  & 70.17 & 44.14 & 60.06 & 47.77 & 21.77 \\
    \hlbox{\ac{ra}}ResNet-101     & 46M  & 71.88 & 46.26 & 61.89 & 49.30 & 23.01 \\
    \hlbox{\ac{ra}}\ac{wrn}-101-2 & 104M & 73.44 & 48.94 & 63.49 & 51.03 & 25.31 \\
\midrule
    PoolFormer-M12 \cite{debenedetti2022light}        & 22M & 66.16 & 34.72 & - & - & -\\ 
    DeiT-S \cite{bai2021transformers}                 & 22M & 66.50 & 35.50 & - & 40.32 & - \\
    ResNet50 + GELU \cite{bai2021transformers}        & 26M & 67.38 & 35.51 & 40.27 & - & - \\
    ResNet50 + DiffPure \cite{nie2022diffusion}       & 26M & 67.79 & 40.39 & - & - & - \\
    XCiT-S12 \cite{debenedetti2022light}              & 26M & 72.34 & 41.78 & - & - & - \\
    ConvNeXt-T \cite{singh2023revisiting}             & 29M & 67.60 & 41.60 & - & - & -\\ 
    XCiT-M12 \cite{debenedetti2022light}              & 46M & 74.04 & 45.24 & - & - & - \\
    \acs{wrn}-50-2 \cite{salman2020adversarially}     & 69M & 68.41 & 38.14 & 55.86 & 41.24 & 16.29 \\
    \acs{wrn}-50-2 + DiffPure \cite{nie2022diffusion} & 69M & 71.16 & 44.39 & - & - & - \\
    Vit-B/16 \cite{mo2022adversarial}                 & 86M & 69.10 & 34.62 & - & 37.52 & - \\
    Swin-B \cite{mo2022adversarial}                   & 88M & 74.36 & 38.61 & - & 40.87 & - \\
    XCiT-L12 \cite{debenedetti2022light}             & 104M & 73.76 & 47.60 & - & - & - \\
\bottomrule
\end{tabular}
}
\label{tab:appx-imagenet}
\end{table}

\newpage
\section{Architecture Details}
\label{sec:appx-arch}

Table~\ref{tab:appx-arch} contains the details of all the robustified architectures mentioned in the paper. 
For depth and width, we present the list of total depths and widths in each stage and compute the corresponding \ac{wd} ratio. 
For the stem stage, we use the convolution stem stage with postponed downsampling operation and increase the output channels to 96. 
Regarding the design of the residual block, we append the \ac{se} block ($r = 4$) to the $3 \times 3$ convolution layer and replace \ac{relu} with \ac{silu}.

\begin{table}[!htbp]
\renewcommand{\arraystretch}{0.9}
\small
\centering
\caption{Details of all the robustified architectures mentioned in the paper.}
\scalebox{0.83}{
\begin{tabular}{lrccrccc}
\toprule
   Model & \#Param. & Depth & Width & \acs{wd} ratio & Stem width & \acs{se} & Activation \\
\midrule
    \hlbox{\acs{ra}}ResNet-50 & 26M & [5, 8, 13, 1] & [36, 72, 140, 270] & 8.99 & 96 & $r = 4$ & \acs{silu}\\
    \hlbox{\acs{ra}}\acs{wrn}-22-10 & 27M & [13, 15, 2] & [120, 240, 480] & 12.62 & 96 & $r = 4$ & \acs{silu} \\
    \hlbox{\acs{ra}}\acs{wrn}-28-10 & 37M & [14, 16, 3] & [128, 256, 512] & 12.57 & 96 & $r = 4$ & \acs{silu} \\
    \hlbox{\acs{ra}}ResNet-101 & 46M & [7, 11, 18, 1] & [42, 84, 166, 328] & 7.62 & 96 & $r = 4$ & \acs{silu} \\
    \hlbox{\acs{ra}}\acs{wrn}-34-12 & 67M & [18, 20, 5] & [144, 288, 576] & 11.20 & 96 & $r = 4$ & \acs{silu} \\
    \hlbox{\acs{ra}}\acs{wrn}-101-2 & 104M & [7, 11, 18, 1] & [64, 128, 252, 504] & 11.59 & 96 & $r = 4$ & \acs{silu} \\
    \hlbox{\acs{ra}}\acs{wrn}-70-16 & 267M & [30, 31, 10] & [216, 432, 864] & 10.57 & 96 & $r = 4$ & \acs{silu} \\
\bottomrule
\end{tabular}
}
\label{tab:appx-arch}
\end{table}

\section{Training curves and convergence rate}
\label{sec:appx-training-curve}

\begin{figure}[h]
  \centering
  \includegraphics[width=0.9\linewidth]{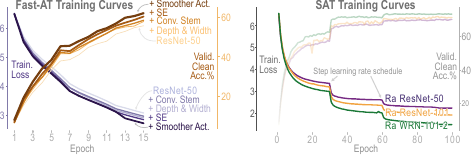}
  \caption{Visualization of the Fast-\ac{at} curves of individual architectural modifications (Table~\ref{tab:roadmap}), and the \ac{sat} curves of the final robustified model (Table~\ref{tab:results-imagenet}).
  We observed that a lower training loss leads to a higher robustness as expected and no catastrophic overfitting occurs during training.}
   \label{fig:training_curve}
\end{figure}

\end{document}